\pdfoutput=1
\documentclass[11pt]{article}

\usepackage[final]{acl}

\usepackage{times}
\usepackage{latexsym}

\usepackage[T1]{fontenc}
\usepackage[utf8]{inputenc}

\usepackage{microtype}

\usepackage{inconsolata}

\usepackage{graphicx}
\usepackage{multirow}
\usepackage{subcaption}
\usepackage{booktabs}
\usepackage{adjustbox}
\usepackage{enumitem}
\usepackage{array, tabularx, bm}
\usepackage{arydshln}
\usepackage{xurl}
\usepackage{amsmath}
\usepackage{pifont}% http://ctan.org/pkg/pifont

\newcolumntype{P}[1]{>{\centering\arraybackslash}p{#1}}
\definecolor{gold}{RGB}{255, 215, 0}    % Gold
\definecolor{silver}{RGB}{192, 192, 192} % Silver
\definecolor{bronze}{RGB}{205, 127, 50}  % Bronze
\newcommand{\gol}{\cellcolor{gold}}
\newcommand{\sil}{\cellcolor{silver}}
\newcommand{\bro}{\cellcolor{bronze}}

\newcommand{\red}[1]{{\color{Red}#1}}
\newcommand{\green}[1]{{\color{Green}#1}}

\newcommand{\ub}[1]{\textbf{\underline{#1}}}

\newcommand{\cmark}{\ding{51}} % checkmark
\newcommand{\xmark}{\ding{55}} % cross

\title{SPHERE: Unveiling Spatial Blind Spots in \\Vision-Language Models Through Hierarchical Evaluation}

\author{
 \textbf{Wenyu Zhang\textsuperscript{1}},
 \textbf{Wei En Ng\textsuperscript{2}},
 \textbf{Lixin Ma\textsuperscript{3*}},
 \textbf{Yuwen Wang\textsuperscript{2*}},
 \textbf{Junqi Zhao\textsuperscript{4*}}, 
 \textbf{Allison Koenecke\textsuperscript{5}} \\
 \textbf{Boyang Li\textsuperscript{4}}, 
 \textbf{Lu Wang\textsuperscript{1}}
\\
 \textsuperscript{1}Institute for Infocomm Research (I$^\text{2}$R), Agency for Science, Technology and Research (A*STAR) \\
 \textsuperscript{2}National University of Singapore (NUS), 
 \textsuperscript{3}Tongji Unversity \\
 \textsuperscript{4}Nanyang Technological University (NTU),
 \textsuperscript{5}Cornell University
\\
 }

\begin{document}
\maketitle
\begin{abstract}
Current vision-language models may grasp basic spatial cues and simple directions (e.g. left, right, front, back), but struggle with the multi-dimensional spatial reasoning necessary for human-like understanding and real-world applications. 
To address this gap, we develop SPHERE (\ub{S}patial \ub{P}erception and \ub{H}ierarchical \ub{E}valuation of \ub{RE}asoning), a hierarchical evaluation framework supported by a new human-annotated dataset. SPHERE systematically probes models across increasing levels of complexity, from fundamental skills to multi-skill integration and high-level reasoning that combines spatial, visual, and logical understanding.
Benchmark evaluation of state-of-the-art models reveals significant deficiencies, especially in reasoning about distance and proximity, understanding both egocentric and allocentric perspectives, and applying spatial logic in physical contexts. These findings expose critical blind spots in existing models and underscore the need for more advanced spatial reasoning techniques, driving the development of vision-language models that align more closely with human spatial cognition. The SPHERE benchmark is available at \href{https://github.com/zwenyu/SPHERE-VLM}{this repository}.
\end{abstract}

\footnotetext{$^*$ Contributed equally to this work; authors are listed in alphabetical order.}

\section{Introduction}

Spatial perception and reasoning play an essential role in how vision-language models (VLMs) understand complex, context-rich environments \cite{du2024embspatialbench, liu2023visualspatial, Zeng2024visualgrounding}. These capabilities are crucial for the deployment of VLMs in physical-world applications such as robotics, embodied AI, or human-assistive systems \citep{Nair2022R3M, brohan2023RT2, stone2023openworldMOO, Leal2023SARART}. 

\begin{figure*}[tb]
    \centering
    \includegraphics[width=0.8\linewidth]{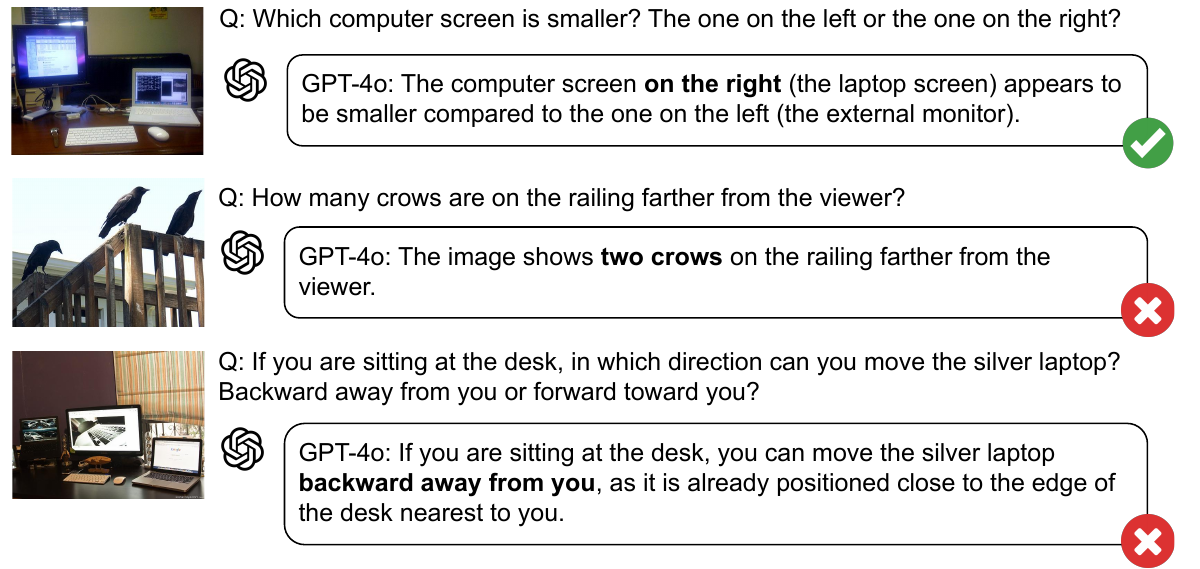}
    \vspace{-4mm}
    \caption{State-of-the-art models such as GPT-4o still has difficulty on questions that require multiple spatial, visual and reasoning skills. 
    % We bolded the words in the first example to highlight the model's answer. 
    GPT-4o itself bolded the words in the second and third examples.}
    \label{fig:gpt_examples}
    \vspace{-2mm}
\end{figure*}

Current research efforts to enhance spatial sensing in VLMs have incorporated features such as depth information, object bounding boxes, and simple spatial directions (e.g. left/right, inside/outside, front/back) into the models \cite{chen2024spatialvlm, zhao2023enhancing, spacellava, cai2024spatialbot, cheng2024spatialrgpt}. 
% These state-of-the-art VLMs can leverage the information to gain better understanding of object positions and relative distances. 
However, the enhancements often focus on isolated, simplistic spatial cues, and therefore limit the models' ability to handle the complexities of real-world scenarios. Practical applications demand more sophisticated spatial perception and reasoning capabilities, akin to human-like understanding. Figure~\ref{fig:gpt_examples} demonstrates failure cases of GPT-4o. In question 2, GPT-4o has difficulty counting while interpreting object proximity. In question 3, GPT-4o understands that the laptop is at the edge of the desk, but fails to reason that the laptop cannot be moved further back.

\begin{figure*}[tb]
    \centering
    \includegraphics[width=\linewidth]{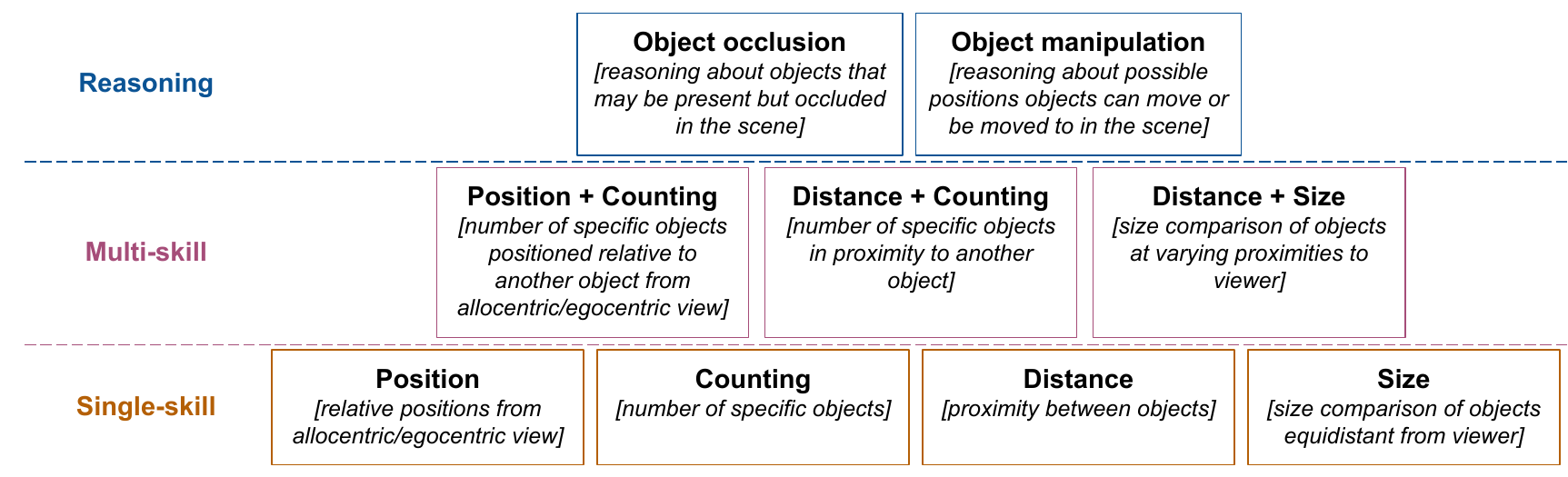}
    \vspace{-4mm}
    \caption{The proposed SPHERE framework evaluates vision-language models on a hierarchy of tasks, advancing from single-skill tasks to multi-skill tasks, and ultimately to complex reasoning tasks that require the integration of multiple spatial and visual cues with logical reasoning abilities.}
    \label{fig:overview_hierarchy}
    \vspace{-2mm}
\end{figure*}

\begin{table}[tb]
\centering
\setlength{\tabcolsep}{2pt}
\begin{adjustbox}{max width=\linewidth}
\begin{tabular}{l*{4}{P{2cm}}}
\toprule[1pt]\midrule[0.3pt]
\textbf{Model} & \multicolumn{1}{c}{Single-skill}  & \multicolumn{1}{c}{Multi-skill} & \multicolumn{1}{c}{Reasoning} & \multicolumn{1}{c}{Overall} \\ \midrule
Human baseline & 95.4 & 92.5 & 89.0 & 93.0 \\
Random baseline$^*$ & 50.0 & 44.3 & 50.0 & 49.1 \\
Best model score  & 78.2 & 58.6 & 64.7 & 67.9 \\
\gol{Rank 1}  & Gemini 2.0 Flash & GPT-4o & GPT-4o & GPT-4o \\
\sil{Rank 2}  & GPT-4o & Qwen2.5-VL & SpatialBot-RGB & Qwen2.5-VL\\
\bro{Rank 3}  & Qwen2.5-VL & LLaVA-OneVision & Llama-3.2-Vision & Gemini 2.0 Flash \\
\midrule[0.3pt]\bottomrule[1pt]
\end{tabular}
\end{adjustbox}
\caption{VLM rankings on SPHERE tasks. $^*$ Random baseline scores excludes open-ended counting tasks. \label{tab: results_summary}}
\vspace{-4mm}
\end{table}

In this work, we propose a new benchmarking framework called SPHERE (\ub{S}patial \ub{P}erception and \ub{H}ierarchical \ub{E}valuation of \ub{RE}easoning), to systematically disentangle and measure spatial skills, in order to pinpoint model strengths and weaknesses and guide future research. We are motivated by prior research findings \cite{West2023paradox, Tiong2024factors} which reveal fundamental differences between machine and human intelligence. While LLMs excel in generating coherent and contextually relevant responses for advanced topics, they often exhibit surprising errors in basic understanding \cite{West2023paradox}. 
We cannot assume good performance in one task to always transfer to a seemingly correlated or even more basic task.

SPHERE, illustrated in Figure~\ref{fig:overview_hierarchy}, is a hierarchical evaluation framework that covers a spectrum of basic and advanced tasks.
We begin with single-skill tasks (i.e. object positions, distances, sizes and counts), then expand to multi-skill tasks that require multiple spatial and visual skills (e.g. counting objects at specific positions, counting objects at certain distances, etc.). We also design two novel reasoning tasks (i.e. reasoning about occluded objects and object manipulation) that require advanced understanding of a scene as a 3-dimensional environment and the objects as physical entities that can be interacted with. We measure reasoning ability as the ability to infer about scenes or objects not explicitly observed but physically plausible in the input images.
% Furthermore, we design a novel study of model ability to infer about scenes or objects not explicitly observed but physically plausible in the input images, through two challenging reasoning tasks (i.e. reasoning about occluded objects and object manipulation). To be successful at the two tasks, models need to interpret multiple spatial and visual cues and to make inferences via logical reasoning. They need to demonstrate advanced understanding of the scene as a 3-dimensional environment and the objects as physical entities that can be interacted with.
These challenging tasks allow us to assess the alignment of VLMs with human-like spatial reasoning capabilities.
Our contributions are summarized as follows:
\begin{itemize}
    \item We design a hierarchical evaluation framework to study the spatial perception and reasoning capabilities of models, progressing from single-skill, multi-skill, to complex reasoning tasks. The reasoning tasks are further decomposed into simpler sub-tasks to disentangle spatial perception and reasoning.
    \item We manually annotate and curate an evaluation dataset using real-world images from MS COCO based on the designed tasks. 
    \item We evaluate state-of-the-art VLMs and reveal significant shortcomings in their capabilities. These models especially lack the ability to understand distance and proximity, to reason from both allocentric and egocentric viewpoints, and to perform complex reasoning in a physical world context.
\end{itemize}

% \cite{yamada2023evaluating} evaluated state-of-the-art LLMs on their spatial capabilities, and find the need to explicitly train the models with spatial tasks to instil specific spatial skills. Since \cite{yamada2023evaluating} used text descriptions alone, the evaluation is constrained to simple patterns and has limited applicability to multimodal inputs.

% robotics application motivation:

% Existing multimodal LLMs typically lack spatial understanding (e.g. object dimensions, distances between objects, spatial relationships such as “between” and “behind”, reasoning about 3D world from visual inputs). These skills are essential for applications such as robotics in manipulation tasks or navigation in physical environments \citep{Nair2022R3M, brohan2023RT2, stone2023openworldMOO, Leal2023SARART}.
\section{Related works}

\subsection{Spatial Capabilities in Multimodal LLMs}

The majority of literature on multimodal LLMs focus on their conversational ability to respond fluently to the queries of human users through natural language \citep{Dai2023InstructBLIP, liu2023llava, liu2024llavanext, lyu2023macaw} or other modalities such as audio and image \cite{wu2023nextgpt, tang2024codi2}.
More recently, works have started to explore the integration of spatial capabilities in LLMs to better understand the physical world. 
\citet{Zheng2024BAT} developed BAT based on text and audio by synthesizing data with a spatial audio simulator, and \citet{devnani2024audio} trained a spatially-aware audio and text embedding model.
SpatialBot \cite{cai2024spatialbot} directly includes pairs of RGB and depth images in existing datasets for training.
A popular approach for vision-language models is to use existing algorithms and models (e.g. object detection, segmentation, depth estimation) to extract additional information from available data, and then incorporate these enhanced data for training or finetuning \citep{chen2024spatialvlm, zhao2023enhancing, spacellava, cai2024spatialbot, cheng2024spatialrgpt}.
The resulting models have improved capabilities to infer spatial relationships between objects and to estimate object dimensions from natural images. 

\subsection{Benchmark Evaluations}

Spatial reasoning is necessary for VLMs to understand and interact with the physical world. This has spurred the development of numerous benchmarks that vary in scope, tasks, and the types of spatial skills they focus on. Some benchmarks \cite{li2024seedbench, yu2024mmvet} involve spatial capabilities in only a small subset of tasks.
EmbSpatial-Bench \cite{du2024embspatialbench} evaluates only a limited number of positional relationships, namely `above', `below', `left', `right', `far' and `close`. 
\citet{Zeng2024visualgrounding} tests the effects of spatial relations on visual grounding tasks.
VSR \cite{liu2023visualspatial} includes proximity and topological concepts. 
SpatialBench \cite{cai2024spatialbot} consists of a small dataset of 120 images to evaluate spatial relationships, counting and size comparison between objects.
Q-Spatial Bench \cite{liao2024qspatial} and SpatialRGPT-Bench \cite{cheng2024spatialrgpt} focuses on model ability to recognize quantitative information such as metric distances and object sizes.
These evaluations and datasets do not disentangle the effects of multiple spatial and/or visual skills. 
CLEVR \cite{Johnson2016CLEVRAD} tests compositional reasoning with multiple spatial and visual skills, but the dataset is synthetic and the objects comprise only simple 3D shapes such as spheres, cubes and cylinders.
\citet{kamath2023whatsup} isolates the effect of spatial attributes by varying the spatial relations of objects while fixing object identities, but the collected dataset focuses only on positional relationships. We study the interaction effect of a wider range of skills, and introduce a novel study of model ability to infer about scenes or objects not explicitly observed but physically plausible in the input images. 
\section{The SPHERE Benchmark}

\subsection{Overview}

The SPHERE evaluation framework is designed to systematically disentangle essential spatial and visual capabilities and to assess VLM performance in tasks that can guide physical world actions. The hierarchy of tasks is illustrated in Figure~\ref{fig:overview_hierarchy}. The first two levels emphasize image understanding, while the final level builds upon image understanding and focuses on analysis and reasoning.

SPHERE begins at the first level with 4 single-skill tasks: Position, Counting, Distance and Size. We choose these skills because they are foundational building blocks for higher-order tasks, and are of interest to the research community as demonstrated through existing efforts to endow VLMs with these skills \cite{cai2024spatialbot, chen2024spatialvlm, du2024embspatialbench}. 
At the second level, the first level skills are integrated into 3 multi-skill tasks: Position + Counting, Distance + Counting, and Distance + Size. We choose to combine 2 skills at a time to minimize confounding factors. Tasks in this category are more challenging. Specifically, Distance + Size require models to understand the concept of size constancy, instead of simply measuring size by the number of pixels occupied in an image.
At the third level, 2 reasoning tasks assess models on their ability to reason about object occlusion and object manipulation. These tasks reflect scenarios commonly encountered in the physical world, such as inferring the presence or position of hidden objects or determining how objects can be moved or rearranged. Logical reasoning together with multiple spatial and visual understanding abilities may be required to successfully answer the questions in this category.

\subsection{Benchmark Dataset}

We manually curate and annotate a question-answering dataset for SPHERE, as automatic generation with VLMs remain unreliable due to accuracy limitations, as demonstrated in Table~\ref{tab: results_summary} and GPT-4o failure cases in Figure~\ref{fig:gpt_examples}. We use images from the test split of MS COCO-2017 \cite{Lin2014MicrosoftCOCO} due to its rich diversity in scenes and object categories.
Since COCO-2017 is a widely-used and publicly available dataset, its images are likely to be in-distribution with respect to training datasets used by VLMs, ensuring that the evaluation aligns with their learned representations. This alignment reduces potential biases stemming from dataset shifts.

\begin{figure}
    \centering
    \includegraphics[width=\linewidth]{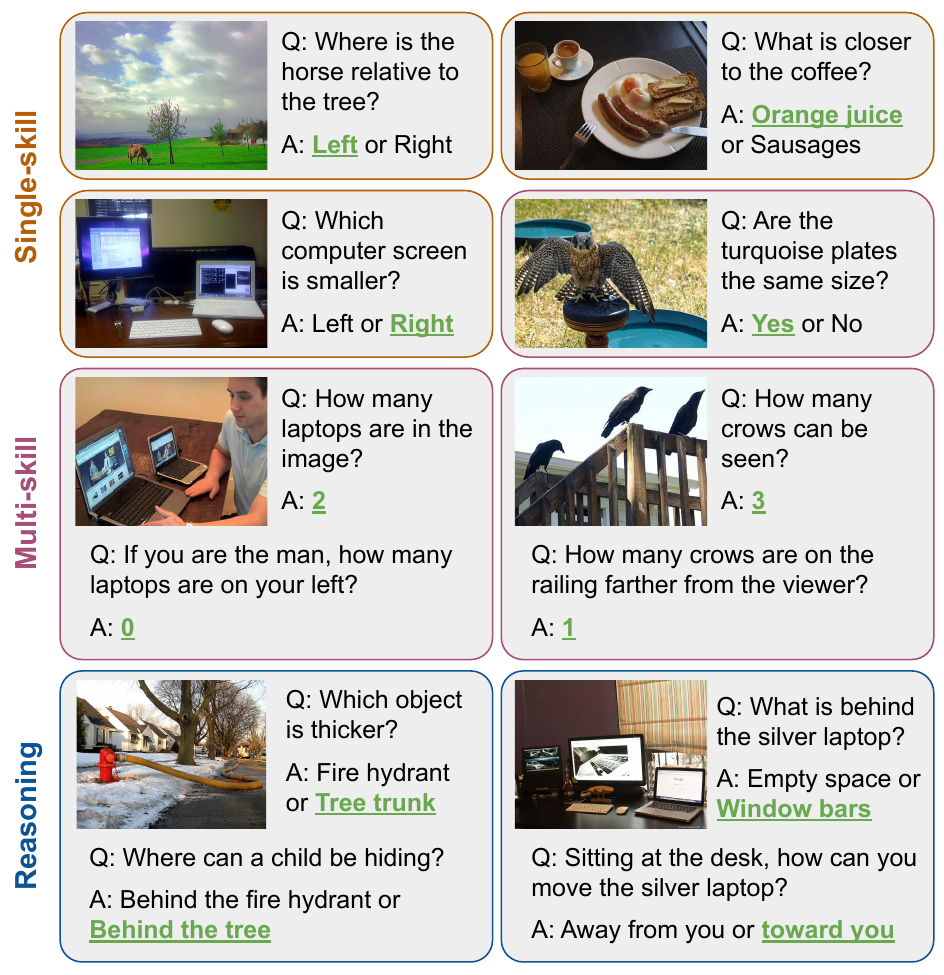}
    \caption{Examples of open-ended questions for counting-related tasks and multiple-choice questions for other tasks in the annotated SPHERE dataset. Ground-truth answers are in {{\color{OliveGreen} \underline{green}}}.}
    \label{fig:annotation_examples}
    \vspace{-4mm}
\end{figure}

Annotations for each task are conducted and cross-verified by at least two authors for consistency and accuracy. Questions that did not elicit the same responses are discarded and replaced. The general guidelines for annotation are:
\begin{itemize}[noitemsep]
    \item Ensure non-ambiguity and diversity of question and answers.
    \item Omit images that may be irrelevant or ambiguous for the task.
    \item Counting-related tasks require open-ended numerical responses, while other tasks use multiple-choice questions (MCQs) with two options by default and three options when additional clarity is needed.
\end{itemize}
We use diverse types of options for MCQs, including boolean (e.g. yes vs. no), position descriptor (e.g. left vs. right), and name of object. Figure~\ref{fig:annotation_examples} shows examples of annotated data.

\begin{figure}
    \centering
    \includegraphics[width=\linewidth]{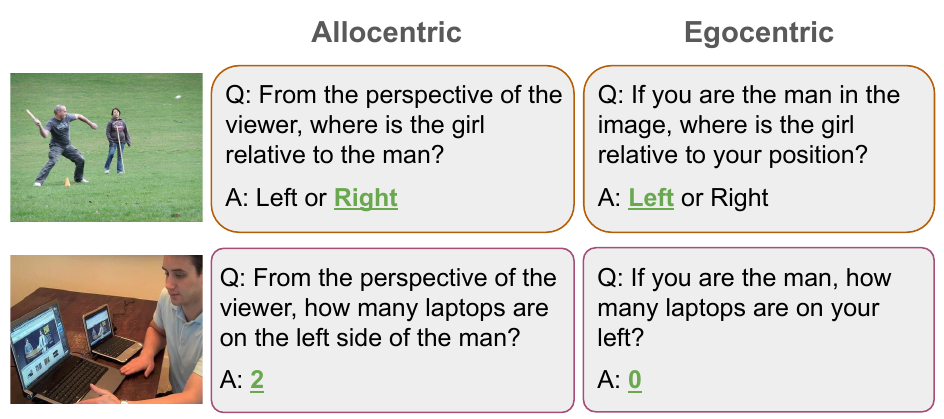}
    \caption{Examples of position-related questions asked from allocentric and egocentric viewpoints.}
    \label{fig:viewpoint_examples}
    \vspace{-2mm}
\end{figure}

\subsubsection{Dataset Design and Statistics}
\label{subsec: dataset design and statistics}

We annotate a total of 2,285 question-answer pairs. The sample size for each task is delineated in Table~\ref{tab: dataset_stats}. Detailed dataset statistics and analysis of dataset diversity are in Appendix \ref{sec:detailed-dataset-statistics}.

\begin{table}[tb]
\centering
\setlength{\tabcolsep}{2pt}
\begin{adjustbox}{max width=0.9\linewidth}
\begin{tabular}{llr}
\toprule[1pt]\midrule[0.3pt]

\textbf{Task level} & \textbf{Task type} & \textbf{Sample size} \\ \hline
\multirow{5}{*}{{Single-skill}} 
& \multirow{2}{*}{Position} & Ego.: 172 \\
& & Allo.: 185\\ \cline{2-3}
& Counting & 201 \\ \cline{2-3}
& Distance & 202 \\ \cline{2-3}
& Size & 198 \\ \hline

\multirow{4}{*}{{Multi-skill}} 
& \multirow{2}{*}{Position + Counting} & Ego.: 64 \\
& & Allo.: 105\\ \cline{2-3}
& Distance + Counting & 158 \\ \cline{2-3}
& Distance + Size & 199 \\ \hline

\multirow{4}{*}{{Reasoning}} 
& \multirow{2}{*}{Object occlusion} & Intermed.: 202 \\
& & Final: 200\\ \cline{2-3}
& \multirow{2}{*}{Object manipulation} & Intermed.: 199 \\
& & Final: 200\\
\midrule[0.3pt]\bottomrule[1pt]
\end{tabular}
\end{adjustbox}
\caption{The annotated dataset has a total of 2,285 question-answer pairs. For position-related tasks, we annotate for both egocentric (ego.) and allocentric (allo.) viewpoints. For reasoning tasks, we additionally annotate for intermediate (intermed.) questions. \label{tab: dataset_stats}}
\vspace{-2mm}
\end{table}

\begin{figure}[tb]
    \centering
    \includegraphics[width=0.7\linewidth]{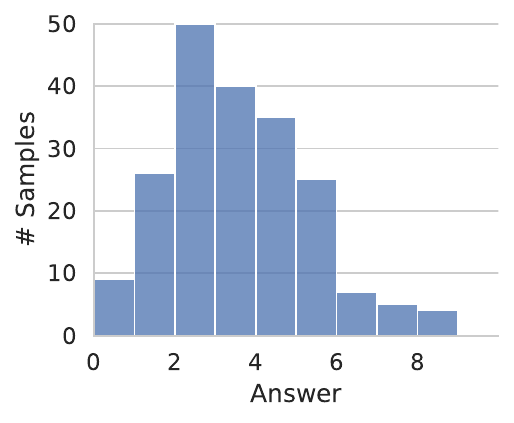}
    \vspace{-4mm}
    \caption{Distribution of ground-truth answers for the counting task.}
    \label{fig:count_distribution}
    \vspace{-4mm}
\end{figure}

\noindent \textbf{Single-skill tasks.}
We annotate for 4 single-tasks. Basic object recognition capabilities are assumed.
\begin{itemize}[noitemsep]
\item \textbf{Position:} We pose questions on relative positional relationships (e.g. left vs. right, in front vs. behind, on top vs. below) between objects in the scene. For entities with clearly defined front and back orientations (e.g. humans, animals, vehicles), we include questions for both \textit{egocentric} and \textit{allocentric} viewpoints.
Egocentric viewpoint questions assess the model's ability to understand spatial relationships from a specified entity's perspective, such as identifying objects to the entity's left or right relative to its orientation.
Allocentric viewpoint questions evaluate the model's ability to describe spatial relationships from the external point of view of the camera or observer. Example questions are shown in Figure~\ref{fig:viewpoint_examples}.

\item \textbf{Counting}: These questions require the model to count the number of specified objects visible in the image. To evaluate the model's susceptibility to object hallucinations, we included `trick' questions where the correct answer is zero. Figure~\ref{fig:count_distribution} illustrates the distribution of ground-truth answers.

\item \textbf{Distance}: We ask questions on the relative proximity (e.g. closer vs. farther) between objects, and between objects and the viewer.

\item \textbf{Size}: We ask questions on the relative size (e.g. smaller vs. bigger, shorter vs. taller) of objects. We focus on objects positioned approximately equidistant from the camera, where object size can be inferred directly from the number of pixels occupied by the object on the image.
\end{itemize}

\begin{figure}[tb]
    \centering
    \includegraphics[width=\linewidth]{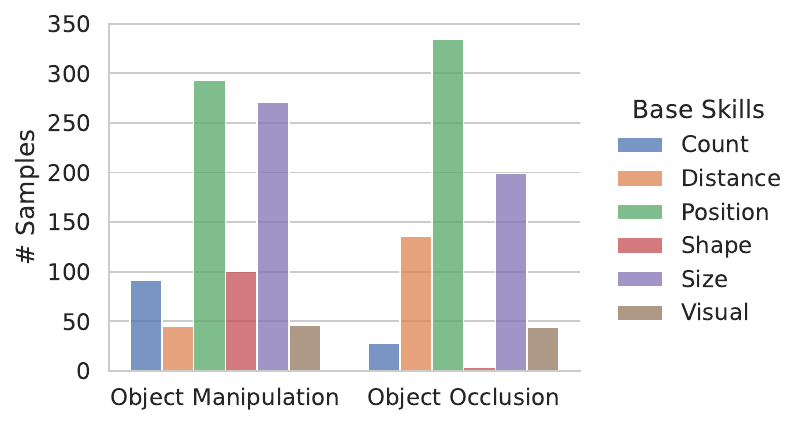}
    \vspace{-4mm}
    \caption{Distribution of base spatial and visual skills required for questions in reasoning tasks.}
    \label{fig:reasoning_skill_distribution}
    \vspace{-4mm}
\end{figure}

\noindent \textbf{Multi-skill tasks:}
We integrate the first level skills into 3 multi-skill tasks.
\begin{itemize}[noitemsep]
\item \textbf{Position + Counting:} Questions require the model to count the number of specified objects that have a specified spatial relationship with another object in the image. We include both egocentric and allocentric viewpoints. Example questions are shown in Figure~\ref{fig:viewpoint_examples}.

\item \textbf{Distance + Counting:} Questions require the model to count the number of specified objects that are at a specified proximity to another object in the image or the observer. An example of a question in Figure~\ref{fig:annotation_examples} is ``How many crows are on the railing farther from the viewer?”

\item \textbf{Distance + Size:} We pose questions about the relative size of objects located at different distances from the camera. To succeed at this task, the model must demonstrate an understanding of size constancy, or the ability to recognize that an object's actual size remains consistent despite changes in its apparent size due to distance or perspective.
\end{itemize}

\noindent \textbf{Reasoning tasks:}
We design 2 tasks to evaluate the model's ability to logically reason about the 3D physical world based on its understanding of the input 2D image. In addition to the final reasoning-based questions, we include understanding-based questions as an intermediate step to check the model's comprehension of the 2D image.
In our design, ``intermediate” questions focus on assessing the perception and understanding of visible objects and scenes. These tasks align with cognitive theories like Piaget's stages of cognitive development, which suggest that perceptual skills are crucial for building more abstract reasoning abilities assessed through the ``final” reasoning questions. The ``final” reasoning questions require the ability to infer the existence, position, or state of partially or non-visible objects and scenes that are physically plausible. These inferences rely on understanding spatial relationships and object permanence, as the missing information could only be revealed by changing viewpoints or interacting with the scene, such as moving objects to uncover hidden ones.

\begin{itemize}[noitemsep]
\item \textbf{Object occlusion:} This task evaluates a model's ability to infer the presence or properties of objects that may be fully occluded from the perspective of observer. From Figure~\ref{fig:annotation_examples}, an example of an intermediate understanding-based question is ``Which object is thicker? Fire hydrant or tree trunk?” The final reasoning-based question is ``Where can a child be hiding? Behind the fire hydrant or behind the tree?” To answer this question correctly, the model needs to reason that a child is more likely to be obscured by the thicker tree.

\item \textbf{Object manipulation:} This task tests a model's ability to reason about potential or hypothetical positions that objects can move to or be moved to within a scene, based on physical constraints. From Figure~\ref{fig:annotation_examples}, an example of an intermediate understanding-based question is ``What is behind the silver laptop? Empty space or window bars?” The final reasoning-based question is ``Sitting at the desk, how can you move the silver laptop? Away from you or toward you?” To answer this question correctly, the model needs to reason that a person at the desk cannot move the laptop further back as the laptop is already positioned against the window bars.
\end{itemize}

Successfully solving reasoning tasks requires multiple spatial and visual skills. We categorize these foundational skills as counting, distance, position, shape, size and other visual skills, such as perceiving color and transparency. We annotate these foundational skills and illustrate the distribution of their occurrences in Figure~\ref{fig:reasoning_skill_distribution}.

\begin{table*}[htb]
\centering
\setlength{\tabcolsep}{2pt}
\begin{adjustbox}{max width=0.95\linewidth}
\begin{tabular}{l*{12}{P{1cm}}P{1cm}}
\toprule[1pt]\midrule[0.3pt]
\textbf{Model} & \multicolumn{5}{c}{Single-skill}  & \multicolumn{4}{c}{Multi-skill} & \multicolumn{3}{c}{Reasoning} \\ \cmidrule(lr){2-6} \cmidrule(lr){7-10} \cmidrule(lr){11-13}
                        & Pos.   & Count. & Dist.  & Size & Avg
                        & P + C   & D + C  & D + S & Avg
                        & Occl. & Manip. & Avg & Overall\\ \midrule
Human baseline          & 90.4 & 98.0 & 98.0 & 95.0 & 95.4 & 95.6 & 91.0 & 91.0 & 92.5 & 87.0 & 91.0 & 89.0 & 93.0 \\
Random baseline$^*$         & 50.0 & - & 50.0 & 50.0 & 50.0 & - & - & 44.3 & 44.3 & 50.0 & 50.0 & 50.0 & 49.1\\ \midrule
& \multicolumn{12}{c}{General purpose VLMs} \\ \midrule
Phi-3.5-Vision (4B)     & 61.8 & 58.7 & 56.7 & 76.0 & 63.3 & 44.4 & 36.1 & 47.2 & 42.6 & 54.4 & 55.5 & 54.9 & 54.3\\
LLaVA-NeXT (7B)         & 53.6 & 68.7 & 54.5 & 70.9 & 61.9 & 42.0 & 35.4 & 36.1 & 37.8 & 48.7 & 49.1 & 48.9 & 50.6 \\
LLaVA-OneVision (7B)    & 60.2 & 76.1 & 64.2 & 84.0 & 71.1 & \sil53.3 & \bro53.8 & 52.9 & \bro53.3 & 53.0 & \bro59.6 & 56.3 & 61.5 \\
Qwen2-VL (7B)           & 63.2 & 74.1 & 53.0 & 78.6 & 67.2 & 36.1 & 33.5 & 50.8 & 40.1 & 54.0 & 55.2 & 54.6 & 55.0\\
Qwen2.5-VL (7B)         & 59.3 & 82.1 & 58.9 & 73.6 & 68.5 & 44.7 & 31.6 & 37.9 & 38.1 & 48.7 & 54.1 & 51.4 & 54.1 \\
Janus-Pro (7B)          & 57.1 & 70.6 & 59.3 & 76.5 & 65.9 & 37.3 & 33.5 & 46.8 & 39.2 & 53.5 & 54.3 & 53.9 & 54.0 \\
InstructBLIP (8B)       & 45.6 & 64.7 & 50.4 & 56.2 & 54.2 & 32.0 & 28.5 & 38.1 & 32.8 & 54.8 & 52.7 & 53.7 & 47.0\\
Idefics2 (8B)           & 50.1 & 45.3 & 49.1 & 58.5 & 50.7 & 22.5 & 27.8 & 40.1 & 30.1 & 51.0 & 53.3 & 52.1 & 44.2\\
InternVL2.5 (8B)        & 62.2 & 72.6 & 64.6 & 78.3 & 69.4 & 50.2 & 43.8 & 41.8 & 45.3 & 52.7 & 53.6 & 53.1 & 57.3\\
Qwen-VL (10B)           & 55.9 & 72.9 & 59.2 & 72.0 & 65.0 & 37.2 & 29.9 & 34.7 & 33.9 & 54.3 & 54.3 & 54.3 & 52.0 \\
Llama-3.2-Vision (11B)  & 58.4 & 53.5 & 52.9 & 67.2 & 58.0 & 29.8 & 27.7 & 40.8 & 32.8 & \bro60.7 & 56.1 & \bro58.4 & 49.7\\
Qwen2-VL (72B)          & 62.8 & \bro80.6 & 60.8 & 85.5 & 72.4 & 37.3 & 38.0 & \gol69.3 & 48.2 & 53.9 & 55.0 & 54.4 & 59.8 \\
Qwen2.5-VL (72B)        & 63.7 & \sil81.1 & \bro69.9 & \sil89.3 & \bro76.0 & \bro51.5 & \sil55.7 & \sil66.4 & \sil57.9 & 50.3 & 56.4 & 53.3 & \sil64.3 \\
Llama-3.2-Vision (90B)  & \bro66.0 & 59.2 & 65.2 & 75.8 & 66.5 & 40.1 & 31.8 & 52.8 & 41.6 & \sil61.6 & 55.3 & \bro58.4 & 56.2\\ 
Gemini 2.0 Flash        & \sil68.6 & \gol82.1 & \gol75.2 & \bro86.9 & \gol78.2 & 50.3 & 52.5 & 49.7 & 50.9 & 48.0 & 49.2 & 48.6 & \bro61.7\\
GPT-4o                  & \gol71.7 & 76.6 & \sil71.3 & \gol89.4 & \sil77.3 & \gol54.4 & \gol62.7 & \bro58.8 & \gol58.6 & \gol66.0 & \gol63.3 & \gol64.7 & \gol67.9 \\ \midrule
& \multicolumn{12}{c}{Spatial VLMs} \\ \midrule
SpatialBot-RGB (3B)     & 55.9 & 67.7 & 52.8 & 73.0 & 62.3 & 38.5 & 39.2 & 38.7 & 38.8 & 57.6 & \sil60.3 & \sil59.0 & 53.6\\
SpatialBot (3B)         & 55.5 & 72.1 & 51.9 & 75.4 & 63.7 & 39.1 & 36.1 & 38.9 & 38.0 & 56.5 & 54.6 & 55.5 & 53.1\\
SpaceMantis (8B)        & 52.7 & 52.7 & 57.0 & 63.1 & 56.4 & 28.4 & 43.7 & 45.6 & 39.2 & 56.6 & 54.2 & 55.4 & 50.4\\
SpatialRGPT-RGB (8B)    & 59.3 & 70.1 & 59.2 & 74.6 & 65.8 & 42.6 & 46.2 & 40.9 & 43.2 & 54.5 & 53.2 & 53.8 & 55.3\\
\midrule[0.3pt]\bottomrule[1pt]
\end{tabular}
\end{adjustbox}
\caption{Average accuracy (\%) of VLMs on SPHERE tasks. Values highlighted in \colorbox{gold}{gold}, \colorbox{silver}{silver} and \colorbox{bronze}{bronze} denote the first, second and third place for each task. The random baseline accuracy for multiple-choice questions is the probability of selecting the correct answer from given options purely by random guessing. $^*$ Average scores for random baseline excludes open-ended counting tasks. \label{tab: results_accuracy}}
\vspace{-2mm}
\end{table*}
\begin{table}[htb]
\centering
\setlength{\tabcolsep}{2pt}
\begin{adjustbox}{max width=\linewidth}
\begin{tabular}{l*{4}{P{1.3cm}}}
\toprule[1pt]\midrule[0.3pt]
\textbf{Model} & \multicolumn{2}{c}{Position}  & \multicolumn{2}{c}{Position + Counting} \\ \cmidrule(lr){2-3} \cmidrule(lr){4-5}
                        & Allo. & Ego. & Allo. & Ego.\\ \midrule
& \multicolumn{4}{c}{General purpose VLMs} \\ \midrule
Phi-3.5-Vision (4B)     & 77.9 & 44.5 & 53.3 & 29.7\\
LLaVA-NeXT (7B)         & 69.3 & 36.7 & 47.6 & 32.8\\
LLaVA-OneVision (7B)    & 73.7 & 45.7 & \gol62.9 & 37.5\\
Qwen2-VL (7B)           & 69.1 & \bro56.0 & 37.1 & 37.5\\
Qwen2.5-VL (7B)         & 69.0 & 48.8 & 49.1 & 37.5 \\
Janus-Pro (7B)          & 65.0 & 48.6 & 37.1 & 37.5 \\
InstructBLIP (8B)       & 43.0 & 48.4 & 31.4 & 32.8\\
Idefics2 (8B)           & 48.5 & 51.9 & 20.0 & 26.6\\
InternVL2.5 (8B)        & 75.7 & 47.7 & \sil59.8 & 34.4\\
Qwen-VL (10B)           & 65.6 & 45.5 & 41.5 & 30.0\\
Llama-3.2-Vision (11B)  & 60.5 & \bro56.0 & 31.6 & 26.9\\
Qwen2-VL (72B)          & 70.3 & 55.6 & 36.2 & 35.9 \\ 
Qwen2.5-VL (72B)        & \bro78.7 & 47.6 & 58.1 & \sil40.6 \\
Llama-3.2-Vision (90B)  & 72.9 & \sil58.6 & 41.1 & \bro38.4\\ 
Gemini 2.0 Flash        & \gol83.2 & 52.9 & \bro55.2 & \gol42.2\\
GPT-4o                  & \sil79.5 & \gol63.4 & \gol62.9 & \sil40.6\\ \midrule
& \multicolumn{4}{c}{Spatial VLMs} \\ \midrule
SpatialBot-RGB (3B)     & 68.6 & 42.2 & 47.6 & 23.4\\
SpatialBot (3B)         & 69.7 & 40.2 & 46.7 & 26.6\\
SpaceMantis (8B)        & 56.2 & 48.8 & 31.4 & 23.4\\
SpatialRGPT-RGB (8B)    & 73.2 & 44.4 & 48.6 & 32.8\\
\midrule[0.3pt]\bottomrule[1pt]
\end{tabular}
\end{adjustbox}
\caption{Average accuracy (\%) of VLMs on position-related questions asked from allocentric (allo.) and egocentric (ego.) viewpoints. Values highlighted in \colorbox{gold}{gold}, \colorbox{silver}{silver} and \colorbox{bronze}{bronze} denote the first, second and third place. \label{tab: viewpoint_accuracy}}
\vspace{-2mm}
\end{table}
\begin{table*}[htb]
\centering
\setlength{\tabcolsep}{2pt}
\begin{adjustbox}{max width=0.9\linewidth}
\begin{tabular}{l*{2}{P{1.3cm}}P{3.5cm}*{2}{P{1.2cm}}P{4cm}}
\toprule[1pt]\midrule[0.3pt]
\textbf{Model} & \multicolumn{3}{c}{Object occlusion}  & \multicolumn{3}{c}{Object manipulation} \\ \cmidrule(lr){2-4} \cmidrule(lr){5-7}
                        & Intermed. & Final & Final (w/ gt intermed.) 
                        & Intermed. & Final & Final (w/ gt intermed.) \\ \midrule
& \multicolumn{6}{c}{General purpose VLMs} \\ \midrule
Phi-3.5-Vision (4B)     & 60.6 & 54.4 & 67.0 $\green{(\uparrow 12.6 )}$ & 59.8 & 55.5 & 57.6 $\green{(\uparrow 2.1 )}$\\
LLaVA-NeXT (7B)         & 60.4 & 48.7 & 54.4 $\green{(\uparrow 5.7 )}$ & 58.5 & 49.1 & 48.8 $\red{(\downarrow 0.3 )}$\\
LLaVA-OneVision (7B)    & 63.9 & 53.0 & 59.8 $\green{(\uparrow 6.8 )}$ & 64.2 & 59.6 & 57.1 $\red{(\downarrow 2.5 )}$\\
Qwen2-VL (7B)           & 63.4 & 54.0 & 60.8 $\green{(\uparrow 6.8 )}$ & 55.1 & 55.2 & 55.4 $\green{(\uparrow 0.2 )}$\\
Qwen2.5-VL (7B)         & 65.2 & 48.7 & 51.1 $\green{(\uparrow 2.4 )}$ & 61.1 & 54.1 & 55.4 $\green{(\uparrow 1.3 )}$ \\
Janus-Pro (7B)          & 58.4 & 53.5 & 55.0 $\green{(\uparrow 1.5 )}$ & 62.6 & 54.3 & 53.0 $\red{(\downarrow 1.3 )}$\\
InstructBLIP (8B)       & 48.7 & 54.8 & 57.8 $\green{(\uparrow 3.0 )}$ & 52.9 & 52.7 & 45.7 $\red{(\downarrow 7.0 )}$\\
Idefics2 (8B)           & 56.5 & 51.0 & 60.0 $\green{(\uparrow 9.0 )}$ & 51.1 & 53.3 & 60.7 $\green{(\uparrow 7.4 )}$\\
InternVL2.5 (8B)        & 63.7 & 52.7 & 58.9 $\green{(\uparrow 6.2 )}$ & 62.9 & 53.6 & 57.2 $\green{(\uparrow 3.6 )}$\\
Qwen-VL (10B)           & 57.8 & 54.3 & 57.3 $\green{(\uparrow 3.0 )}$ & 54.5 & 54.3 & 49.0 $\red{(\downarrow 5.3 )}$\\
Llama-3.2-Vision (11B)  & 52.3 & 60.7 & 62.5 $\green{(\uparrow 1.8 )}$ & 53.7 & 56.1 & 57.5 $\green{(\uparrow 1.4 )}$\\
Qwen2-VL (72B)          & 65.6 & 53.9 & 62.4 $\green{(\uparrow 8.5 )}$ & 68.5 & 55.0 & 76.9 $\green{(\uparrow 21.9 )}$\\ 
Qwen2.5-VL (72B)        & 67.3 & 50.3 & 57.6 $\green{(\uparrow 7.3 )}$ & 64.2 & 56.4 & 66.9 $\green{(\uparrow 10.5 )}$\\
Llama-3.2-Vision (90B)  & 62.7 & 61.6 & 68.0 $\green{(\uparrow 6.4 )}$ & 59.8 & 55.3 & 74.0 $\green{(\uparrow 18.7 )}$\\ 
Gemini 2.0 Flash        & 62.4 & 48.0 & 61.0 $\green{(\uparrow 13.0 )}$ & 65.5 & 49.2 & 58.3 $\green{(\uparrow 9.1 )}$\\
GPT-4o                  & 67.3 & 66.0 & 72.0 $\green{(\uparrow 6.0 )}$ & 68.5 & 63.3 & 76.4 $\green{(\uparrow 13.1 )}$\\ \midrule
& \multicolumn{6}{c}{Spatial VLMs} \\ \midrule
SpatialBot-RGB (3B)     & 59.9 & 57.6 & 65.9 $\green{(\uparrow 8.3 )}$ & 60.8 & 60.3 & 54.0 $\red{(\downarrow 6.3 )}$\\
SpatialBot (3B)         & 58.6 & 56.5 & 63.2 $\green{(\uparrow 6.7 )}$ & 59.2 & 54.6 & 56.1 $\green{(\uparrow 1.5 )}$\\
SpaceMantis (8B)        & 63.1 & 56.6 & 62.4 $\green{(\uparrow 5.8 )}$ & 60.1 & 54.2 & 57.6 $\green{(\uparrow 3.4 )}$\\
SpatialRGPT-RGB (8B)    & 61.1 & 54.5 & 63.2 $\green{(\uparrow 8.7 )}$ & 60.1 & 53.2 & 58.5 $\green{(\uparrow 5.3 )}$\\
\midrule[0.3pt]\bottomrule[1pt]
\end{tabular}
\end{adjustbox}
\caption{Average accuracy (\%) of VLMs on the intermediate (intermed.) and final questions in the reasoning tasks. The approach `final (w/ gt intermed.)' denotes that the intermediate questions and ground-truth answers are used as additional information to assist the answering of the final questions.
\label{tab: reasoning_accuracy}}
\vspace{-2mm}
\end{table*}

\section{Evaluations}

\subsection{Setup}

We describe the experimental setup used to evaluate the performance of state-of-the-art VLMs on SPHERE tasks. All experiments were performed on NVIDIA GeForce RTX 3090 GPUs.

\noindent \textbf{Models:} Our evaluation encompasses two model categories: general-purpose VLMs and VLMs with enhanced spatial capabilities.
The general-purpose VLMs evaluated are Phi-3.5-Vision \cite{Abdin2024Phi3TR}, LLaVA-NeXT \cite{liu2024llavanext}, LLaVA-OneVision \cite{li2024llavaonevision}, InstructBLIP \cite{Dai2023InstructBLIP}, Idefics2 \cite{idefics2}, InternVL2.5 \cite{chen2024internvl}, Janus-Pro \cite{chen2025janus}, Qwen-VL \cite{bai2023qwenvl}, Qwen2-VL \cite{Qwen2VL}, Qwen2.5-VL \cite{Qwen2.5-VL}, Llama-3.2-Vision \cite{llama3.2_vision}, Gemini 2.0 Flash \cite{gemini2.0_flash}, and GPT-4o \cite{gpt4o}. The VLMs specifically trained for spatial understanding and reasoning are SpatialBot \citep{cai2024spatialbot}, SpaceMantis \cite{spacemantis}, and SpatialRGPT \cite{cheng2024spatialrgpt}. 
For SpatialRGPT, we use the RGB-only version as the model requires highly accurate mask proposals to make use of depth maps.
% Besides the default version of SpatialBot that accepts both RGB and depth map as inputs, we also include the RGB-only version of SpatialBot as SpatialBot-RGB. 
% SpaceMantis \cite{spacemantis} is fine-tuned from Mantis-8B-siglip-llama3 \cite{spacemantis} following techniques described in \citep{chen2024spatialvlm}. 
All model implementations follow their released versions on Hugging Face. 
We use the default prompt templates and hyperparameters in the demos of the respective models. 

The text instructions given take the form:
\begin{quote}
\textit{``<question> Answer the question directly.”}
\end{quote}
For example, for the first image in Figure~\ref{fig:viewpoint_examples}, the text instruction given is \textit{``From the perspective of the viewer, where is the girl relative to the man? Left or Right? Answer the question directly.”} We added the instruction “Answer the question directly” to reduce the refusal-to-answer rate and to encourage concise answers that align with the options provided. 

\noindent \textbf{Metrics:} We computed both the \textit{validity} and \textit{accuracy} of the VLM responses. Since the VLM generative responses are open-ended, we deem a response as valid if it is a relevant answer to the question, that is, a number is given for a counting question and the response contains one of the options for a multiple-choice question (MCQ). Accuracy measures the correctness of the model responses, and invalid responses are treated as incorrect. For MCQs, we set the baseline accuracy for each question as $1/{\# options}$, since this is the probability of selecting the correct answer purely by random guessing.

To account for the probabilistic nature of VLM generations, we evaluate each question 5 times using different random seeds and average the validity and accuracy scores across these runs. For MCQs, we introduce additional randomness by shuffling the order of the answer options at each seed. This prevents the models from exploiting positional biases in the option sequence and ensures that their performance reflects true understanding rather than reliance on specific patterns. Through these considerations, we increase the robustness of our evaluation, providing a more reliable measure of the models’ generalization and consistency. 

\subsection{Results}

Table~\ref{tab: results_accuracy} shows model accuracy on each of the SPHERE tasks. The results highlight significant room for improvement, especially for multi-skill and reasoning tasks.

The proprietary model GPT-4o outperforms the open-weights models evaluated in almost all tasks and has an overall accuracy of 67.9\%. Qwen2.5-VL (72B) has the second highest overall accuracy of 64.3\%. Interestingly, amongst open-weights models, we observe that the smaller model LLaVA-OneVision (7B) can outperform larger models in terms of overall accuracy, and performs on par with Gemini 2.0 Flash.

For the single-skill tasks, Gemini 2.0 Flash has the highest average accuracy (78.2\%). The best performances on Counting and Size are above 80\%. The best performances on Position and Distance are between 70\% and 80\%, indicating that current VLMs still have difficulty in these two skills. In Section~\ref{subsec: further analysis}, we analyze how viewpoint affects model performance on understanding positions.

For the multi-skill tasks, GPT-4o has the highest average accuracy (58.6\%). The multi-skill tasks are markedly more challenging than the single-skill tasks. In particular, we observe that half of the models perform worse than baseline random guessing on Distance + Size. Qwen2-VL (72B) obtains the highest 69.3\% on Distance + Size, and 85.5\% on Size. 
This gap reflects the reliance of the VLMs on the 2D representation of objects in an image, and measures object size as being directly proportional to the number of pixels occupied. The failure to incorporate the principle of size constancy may cause VLMs to struggle in 3D spaces.

For the reasoning tasks, GPT-4o has the highest average accuracy (64.7\%). Most models perform close to random guessing, suggesting that their training datasets may lack sufficient examples emphasizing reasoning skills. Notably, Gemini 2.0 Flash has the worst average accuracy (48.6\%), below the random guessing baseline.

In general, we note that the spatial VLMs evaluated performed comparably with or better than most similarly-sized general-purpose VLMs, except for LLaVA-OneVision. In particular, SpatialBot (3B) performed comparably with or better than larger models with up to 11B parameters. This shows that the techniques in spatial VLMs can benefit spatial capabilities. Possible reasons on why the spatial VLMs do not excel on the SPHERE tasks may be the lack of training on the specific skills emphasized by the SPHERE tasks, and the lack of model generalization to the SPHERE tasks.

We refer readers to Table~\ref{tab: results_validity} in the Appendix \ref{sec: detailed evaluation results} for detailed validity scores of the models. The validity scores reflect the instruction-following capabilities of the VLMs. Idefics2 has above 95\% validity, and all other models have above 98\% validity.
% Most models effectively follow instructions and provide valid responses, such as numerical answers for counting-related questions and selecting an option from the provided list for multiple-choice questions.
% Common types of invalid responses include providing an image caption instead of addressing the question, refusing to answer, or offering an answer that is not among the given options.

\subsection{Further Analysis}
\label{subsec: further analysis}

\noindent \textbf{Bias toward allocentric or egocentric viewpoint.} We further analyze model performance on position-related tasks by distinguishing between allocentric and egocentric viewpoints. This analysis provides insights into whether VLMs can comprehend object orientations and reason from both relative (object-to-object) and absolute (viewer-to-object) perspectives. Example questions are shown in Figure~\ref{fig:viewpoint_examples}. From the results in Table~\ref{tab: viewpoint_accuracy}, we observe that most VLMs demonstrate a significant bias toward the allocentric viewpoint, with performance gaps as large as $33.4\%$ for Phi-3.5-Vision in the Position task, and $25.4\%$ for LLaVA-OneVision in the Position + Counting task.

\noindent \textbf{Effectiveness of additional information for reasoning.} Since the reasoning tasks first require image understanding of the factual information presented in the images and then logical reasoning to conduct analysis and inference, we further disentagle the two types of abilities.
We measure the strength of model reasoning capabilities alone by directly providing understanding-based information in the input prompts. That is, we prefix the final step question with the intermediate step question and ground-truth answer as:
\begin{quote}
\textit{``Given that for the question: <intermediate step question> The answer is: <intermediate step answer>. <final step question> Answer the question directly.”}
\end{quote}
For the example in Figure~\ref{fig:annotation_examples}, the prompt will be \textit{``Given that for the question: Which object is thicker? Fire hydrant or tree trunk? The answer is: Tree trunk. Where can a child be hiding? Behind the fire hydrant or behind the tree? Answer the question directly.”}

From Table~\ref{tab: reasoning_accuracy}, we observe that incorporating additional information improves final question accuracy across all models for the object occlusion task, but yields mixed results for the object manipulation task. Although the additional information can introduce relevant facts that the models might otherwise overlook, the models sometimes struggle to establish a clear connection between these facts and the specific reasoning they need to conduct.

\noindent \textbf{Skill composition.} 
The different performance of the evaluated models across tasks reflect that they have different compositions of skills, and that proficiency in one task does not necessarily translate to proficiency in related tasks. For instance, relatively weaker basic skills do not necessarily imply correspondingly weaker higher-order skills. In Table~\ref{tab: results_accuracy}, SpaceMantis has 7.3\% lower average accuracy than SpatialBot on the single-skill tasks, but has 1.2\% higher average accuracy than SpatialBot on the multi-skill tasks. In Table~\ref{tab: reasoning_accuracy}, on the object manipulation task, Idefics2 and Llama-3.2-Vision (11B) have attained higher accuracy on the final reasoning questions (53.3\% and 56.1\%) than on the supposedly easier intermediate perception or understanding questions (51.1\% and 53.7\%).

\section{Conclusion}

This work introduces SPHERE, a comprehensive evaluation framework designed to address the critical gaps in spatial understanding and reasoning exhibited by current vision-language models. Through a hierarchy of tasks and a meticulously curated dataset, SPHERE highlights significant limitations in the ability of state-of-the-art models, especially in processing spatial cues such as distance, reasoning from both allocentric and egocentric viewpoints, and analyzing the depicted scene in a physical world context. Through this work, we seek to lay the groundwork for advancing vision-language models toward more human-like spatial perception and reasoning, and to set a benchmark for future research in this domain.

\section{Limitations}

As we seek to ensure the quality of the SPHERE dataset through manual annotations, the dataset may be limited in scope. The framework and the curated tasks may not encompass all possible spatial understanding and reasoning challenges encountered in practice. In addition, the dataset focuses on static images, which may not fully capture the challenges of dynamic spatial understanding and reasoning in real-world scenarios. As VLMs advance in spatial perception and reasoning, automatic annotation methods may become more reliable and warrant exploration in future work.

\section{Ethics Statement}

All real-world images used in our annotated dataset are sourced from the MS COCO-2017 dataset \cite{Lin2014MicrosoftCOCO}, a widely used and publicly available resource. Our annotations focus solely on the spatial properties of the entities in the images and do not involve any personally identifiable information. The authors have unanimously agreed that the annotated data will be used exclusively for academic research purposes, in compliance with the dataset’s terms of use \cite{cocoterms}.

\section*{Acknowledgements}

We thank Chanwoo Hwang from Hanyang University for valuable discussions.
This research is partially supported by the Nanyang Associate Professorship and the National Research Foundation Fellowship (NRF-NRFF13-2021-0006), Singapore.

\bibliography{spatial}

\clearpage

\appendix

\section{Detailed Dataset Statistics}
\label{sec:detailed-dataset-statistics}

We provide additional details of our annotated SPHERE dataset.

\paragraph{Number of options.}

\begin{figure}[tb]
    \centering
    \includegraphics[width=0.75\linewidth]{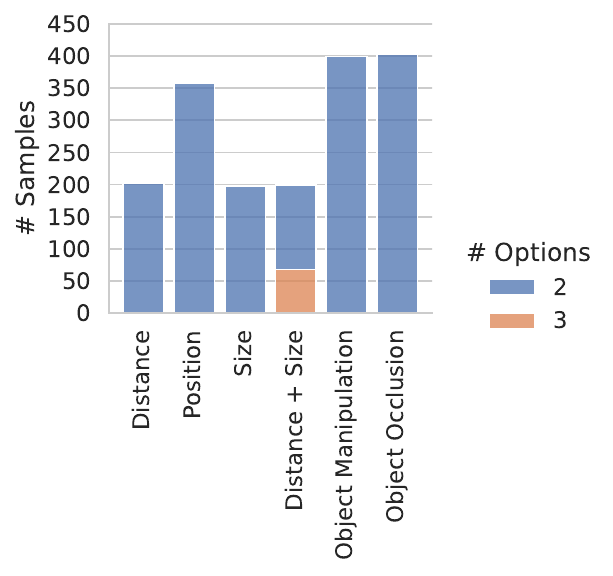}
    \caption{Distribution of number of options for MCQs across all non-counting tasks.}
    \label{fig:num_options_distribution}
\end{figure}

For non-counting tasks, questions are posed as multiple-choice questions (MCQs). Figure \ref{fig:num_options_distribution} shows the distribution of number of options across all tasks. All non-counting tasks except Distance + Size only have two-option MCQs, while around a third of MCQs in Distance + Size have three options. When comparing the size of two objects and asking the models to identify the smaller/bigger object, we provide options of the form `the closer object', `the farther object' and `the two objects are of similar size'.
Such three-option MCQs are designed to evaluate if the model has an understanding of size constancy when similar objects may appear to be of different sizes in the image due to their positioning in the 3D scene.

\paragraph{Numerical answers.}

\begin{figure*}[tb]
    \centering
    \includegraphics[width=0.9\linewidth]{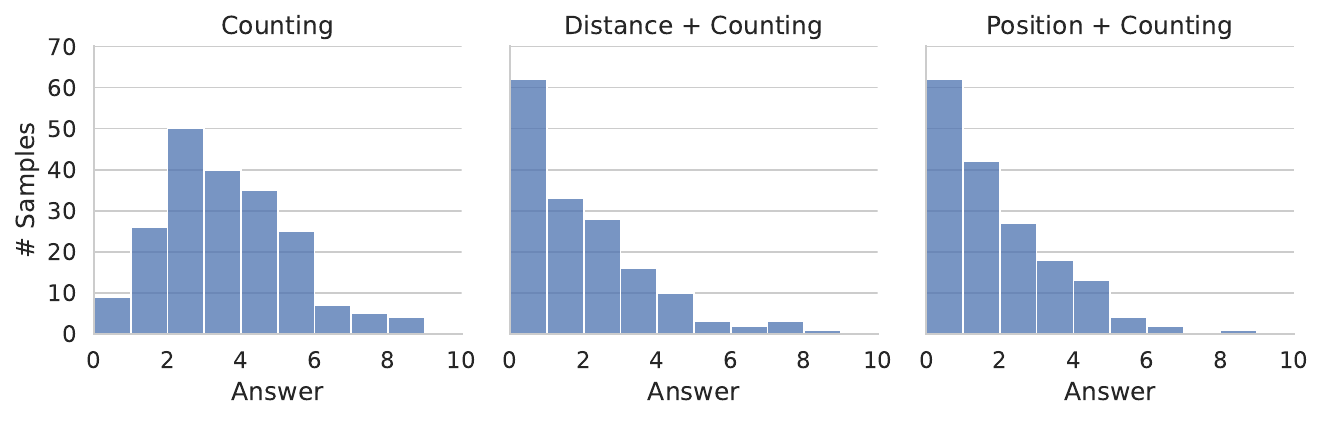}
    \caption{Distribution of ground-truth answers for all counting tasks.}
    \label{fig:count_distribution_full}
\end{figure*}

Counting tasks are phrased as open-ended numerical questions, designed to evaluate the models' ability to recognize objects matching the required criterion while distinguishing such objects as distinct entities. Figure \ref{fig:count_distribution_full} shows the distribution of ground-truth answers across all counting-related tasks. We limit the number of objects to be counted to a maximum of 9 to avoid including overly small objects.

We check the count distributions by fitting a negative binomial model to each distribution using the \texttt{goodfit} function from R's \texttt{vcd} package. The results of Pearson's $\chi^2$ tests, shown in Table \ref{tab: goodness_of_fit}, indicate that all $p$-values exceeded 0.10. This suggests that the count distributions align well with the negative binomial model and are not excessively skewed toward specific values.

\paragraph{Formats.}

\begin{figure*}[tb]
    \centering
    \begin{subfigure}[b]{0.45\textwidth}
        \centering
        \includegraphics[width=\linewidth]{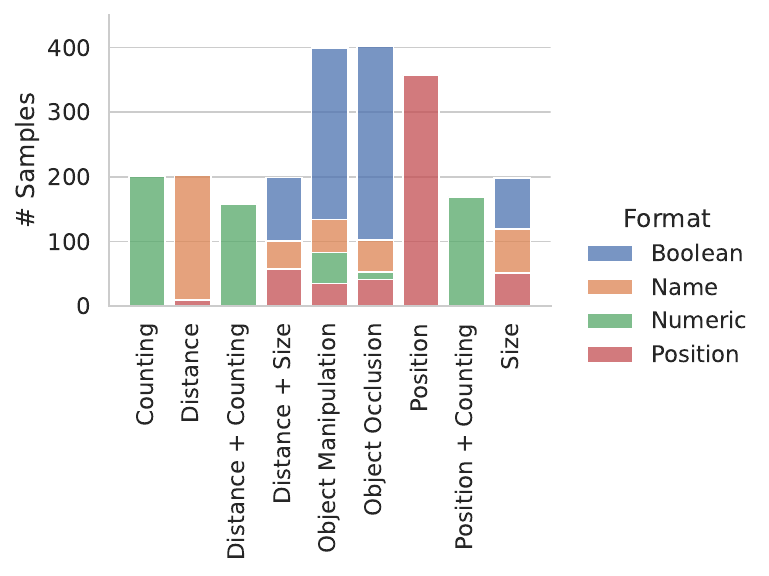}
        \caption{Option formats.}
        \label{fig:format_distribution}
    \end{subfigure}
    \begin{subfigure}[b]{0.45\textwidth}
        \centering
        \includegraphics[width=\linewidth]{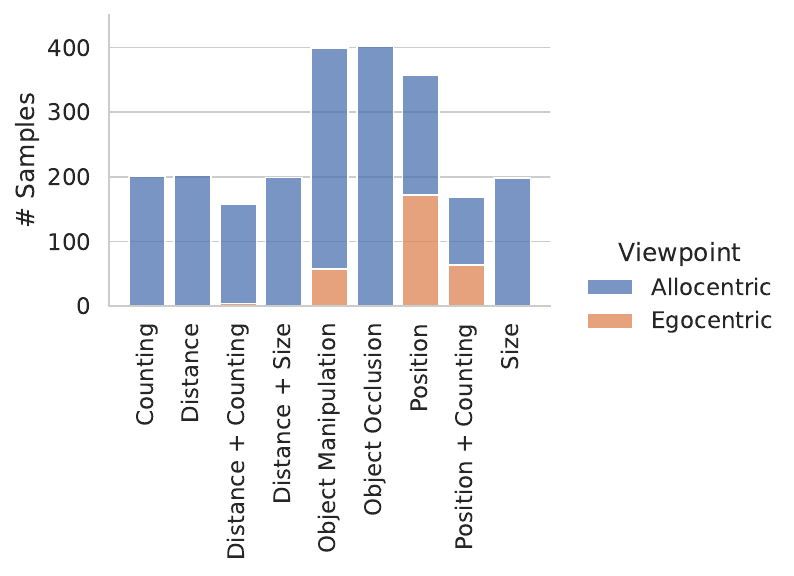}
        \caption{Viewpoints.}
        \label{fig:viewpoint_distribution}
    \end{subfigure}
    \caption{Distribution of option formats and viewpoints across all tasks.}
\end{figure*}

Figure \ref{fig:format_distribution} shows the distribution of the following option formats across all tasks: \begin{itemize}[noitemsep]
\item \textbf{Boolean:} True/false or other similar phrasings, e.g. ``yes'' vs. ``no''.
\item \textbf{Name:} Name of objects or their attributes, e.g. ``empty space'' vs. ``window pane'', ``black laptop'' vs. ``white laptop'', ``curved'' vs ``straight''.
\item \textbf{Numeric:} For counting tasks, a non-negative integer indicating the number of objects matching the required criterion. For non-counting tasks, MCQs may contain numbers in the provided options, e.g. ``1/2'' vs. ``less than 1/4'', or require numerical estimation, e.g. ``less than the length of the painting'' vs. ``more than the length of the painting''.
\item \textbf{Position:} Position of objects, e.g. ``left'' vs. ``right'', ``on top'' vs. ``below''.
\end{itemize}

\paragraph{Viewpoints.}

Figure \ref{fig:viewpoint_distribution} shows the distribution of egocentric vs. allocentric viewpoints across all tasks. Position, Position + Counting, Distance + Counting and Object manipulation comprise both allocentric and egocentric questions, while all other tasks comprise only allocentric questions.

\paragraph{Object concepts.}

\begin{figure*}[tb]
    \centering
    \includegraphics[width=0.9\linewidth]{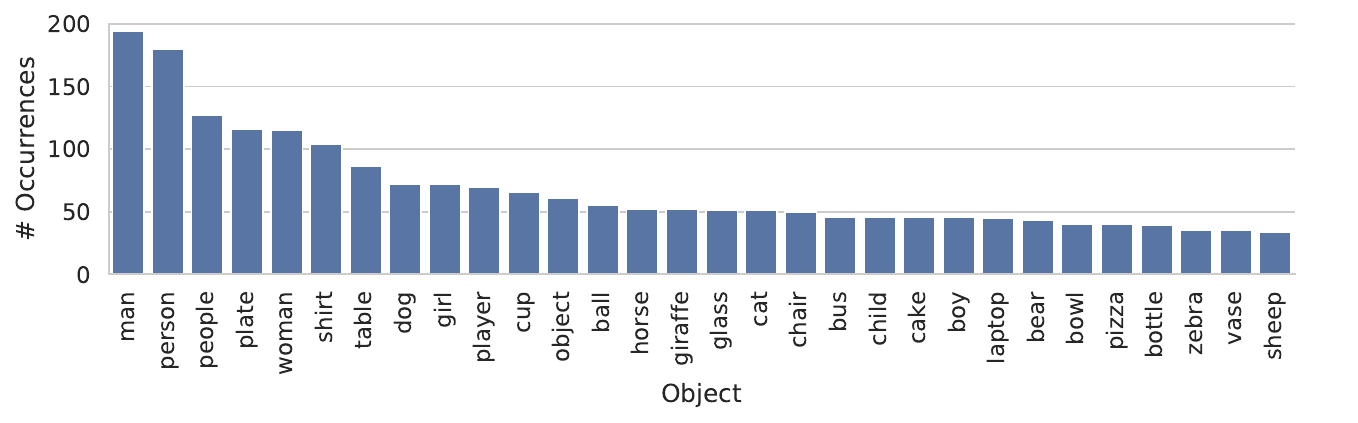}
    \caption{Distribution of object concepts across the dataset.}
    \label{fig:object_distribution}
\end{figure*}

We assess the diversity of object concepts in the annotations by counting the occurrences of unique nouns. Nouns were extracted from each question using the \texttt{en\_core\_web\_sm} model from spaCy, then manually filtered to exclude propositions and commonly referenced objects in question phrasings, such as `camera', `viewer', `photo', `image'. The dataset contains a total of more than 600 object concepts. 
Figure \ref{fig:object_distribution} shows the distribution of object concepts across the dataset for the top 30 concepts.

\begin{table}[tb]
\centering
\begin{tabular}{rlll}
\toprule[1pt]\midrule[0.3pt]

Task & $\chi^2$ & df & $P(> \chi^2)$ \\
\midrule
Counting & 6.2420 & 7 & 0.51179 \\
Distance + Counting & 7.8268 & 6 & 0.25107 \\
Position + Counting & 7.8879 & 5 & 0.16252 \\
\midrule[0.3pt]\bottomrule[1pt]
\end{tabular}
\caption{Goodness-of-fit tests on ground-truth answers for counting tasks fitted to a negative-binomial model.} \label{tab: goodness_of_fit}
\end{table}

%\begin{figure*}[h]
%    \centering
%    \includegraphics[width=0.3\linewidth]{figures/appendix/rootogram_counting_only.png}
%    \includegraphics[width=0.3\linewidth]{figures/appendix/rootogram_distance_and_counting.png}
%    \includegraphics[width=0.3\linewidth]{figures/appendix/rootogram_position_and_counting.png}
%    \caption{Hanging rootogram of ground-truth answers fitted to a negative-binomial model, for all counting tasks. The figures were produced using \texttt{rootogram} from R's \texttt{vcd} package.}
%    \label{fig:rootogram}
%\end{figure*}

\section{Detailed Evaluation Results}
\label{sec: detailed evaluation results}

Table~\ref{tab: results_validity} shows the validity scores of the models across all tasks. These scores reflect the instruction-following capabilities of the VLMs. Most models effectively follow instructions and provide valid responses, such as numerical answers for counting-related questions and selecting an option from the provided list for multiple-choice questions. Idefics2 has above 95\% validity, and all other models have above 98\% validity. Common types of invalid responses include providing an image caption instead of addressing the question, refusing to answer, or offering an answer that is not among the given options.

We provide example questions and answers for each task in Table~\ref{tab: single skill examples}, \ref{tab: multi skill examples} and \ref{tab: reasoning examples}.

\begin{table*}[tb]
\centering
\setlength{\tabcolsep}{2pt}
\begin{adjustbox}{max width=\linewidth}
\begin{tabular}{l*{12}{P{1cm}}P{1cm}}
\toprule[1pt]\midrule[0.3pt]
\textbf{Model} & \multicolumn{5}{c}{Single-skill}  & \multicolumn{4}{c}{Multi-skill} & \multicolumn{3}{c}{Reasoning} \\ \cmidrule(lr){2-6} \cmidrule(lr){7-10} \cmidrule(lr){11-13}
                        & Pos.   & Count. & Dist.  & Size & Avg
                        & P + C   & D + C  & D + S & Avg
                        & Occl. & Manip. & Avg & Overall\\ \midrule
& \multicolumn{12}{c}{General purpose VLMs} \\ \midrule
Phi-3.5-Vision (4B)     & 100.0 & 100.0 & 99.7 & 100.0 & 99.9 & 100.0 & 100.0 & 99.0 & 99.7 & 100.0 & 100.0 & 100.0 & 99.9\\
LLaVA-NeXT (7B)         & 100.0 & 100.0 & 100.0 & 100.0 & 100.0 & 100.0 & 100.0 & 98.8 & 99.6 & 100.0 & 100.0 & 100.0 & 99.9\\
LLaVA-OneVision (7B)    & 99.7 & 100.0 & 99.8 & 98.4 & 99.5 & 100.0 & 100.0 & 97.3 & 99.1 & 99.7 & 100.0 & 99.8 & 99.4\\
Qwen2-VL (7B)           & 97.4 & 100.0 & 98.3 & 99.1 & 98.7 & 100.0 & 100.0 & 98.5 & 99.5 & 98.5 & 99.5 & 99.0 & 99.0\\
Qwen2.5-VL (7B)         & 94.0 & 100.0 & 98.8 & 98.8 & 97.9 & 100.0 & 99.4 & 96.8 & 98.7 & 98.4 & 97.3 & 97.9 & 98.2  \\
Janus-Pro (7B)          & 100.0 & 100.0 & 100.0 & 100.0 & 100.0 & 100.0 & 100.0 & 99.8 & 99.9 & 99.8 & 100.0 & 99.9 & 100.0\\
InstructBLIP (8B)       & 97.1 & 100.0 & 100.0 & 99.7 & 99.2 & 99.4 & 100.0 & 99.9 & 99.8 & 99.8 & 100.0 & 99.9 & 99.6\\
Idefics2 (8B)           & 92.9 & 93.0 & 96.5 & 91.9 & 93.6 & 99.4 & 96.8 & 91.4 & 95.9 & 98.7 & 96.0 & 97.3 & 95.3\\
InternVL2.5 (8B)        & 98.2 & 100.0 & 99.4 & 98.9 & 99.1 & 100.0 & 100.0 & 99.3 & 99.8 & 99.8 & 100.0 & 99.9 & 99.5\\
Qwen-VL (10B)           & 99.0 & 100.0 & 99.5 & 100.0 & 99.6 & 100.0 & 100.0 & 98.4 & 99.5 & 96.8 & 98.8 & 97.8 & 99.1\\
Llama-3.2-Vision (11B)  & 99.0 & 100.0 & 99.0 & 98.3 & 99.1 & 100.0 & 100.0 & 98.6 & 99.5 & 100.0 & 99.9 & 99.9 & 99.4\\
Qwen2-VL (72B)          & 97.6 & 100.0 & 99.5 & 99.0 & 99.0 & 100.0 & 100.0 & 99.3 & 99.8 & 99.6 & 100.0 & 99.8 & 99.5\\ 
Qwen2.5-VL (72B)        & 99.2 & 100.0 & 100.0 & 100.0 & 99.8 & 100.0 & 99.2 & 99.2 & 99.5 & 99.6 & 99.9 & 99.8 & 99.7 \\
Llama-3.2-Vision (90B)  & 95.9 & 100.0 & 99.5 & 98.3 & 98.4 & 100.0 & 100.0 & 98.3 & 99.4 & 100.0 & 98.8 & 99.4 & 99.0\\ 
Gemini 2.0 Flash        & 99.4 & 100.0 & 100.0 & 100.0 & 99.9 & 100.0 & 100.0 & 98.5 & 99.5 & 100.0 & 100.0 & 100.0 & 99.8\\
GPT-4o                  & 99.2 & 99.5 & 98.5 & 97.0 & 98.5 & 100.0 & 100.0 & 94.0 & 98.0 & 98.5 & 100.0 & 99.2 & 98.5\\ \midrule
& \multicolumn{12}{c}{Spatial VLMs} \\ \midrule
SpatialBot-RGB (3B)     & 99.2 & 99.0 & 99.8 & 98.4 & 99.1 & 100.0 & 100.0 & 99.0 & 99.7 & 100.0 & 99.5 & 99.7 & 99.4\\
SpatialBot (3B)         & 99.2 & 100.0 & 100.0 & 100.0 & 99.8 & 100.0 & 99.4 & 99.3 & 99.6 & 100.0 & 100.0 & 100.0 & 99.8\\
SpaceMantis (8B)        & 99.4 & 100.0 & 100.0 & 99.7 & 99.8 & 100.0 & 100.0 & 98.5 & 99.5 & 100.0 & 100.0 & 100.0 & 99.7\\
SpatialRGPT-RGB (8B)    & 98.9 & 100.0 & 100.0 & 99.1 & 99.5 & 100.0 & 100.0 & 99.2 & 99.7 & 97.9 & 95.9 & 96.9 & 98.9\\
\midrule[0.3pt]\bottomrule[1pt]
\end{tabular}
\end{adjustbox}
\caption{Average validity (\%) of VLM responses on SPHERE tasks. \label{tab: results_validity}}
\vspace{-2mm}
\end{table*}

\begin{table*}[tb]
\centering

%%% Position
\begin{subtable}[t]{\textwidth}
\begin{adjustbox}{max width=\textwidth}
\begin{tabular}{m{3cm} m{6cm} m{6cm}}
\toprule
\textbf{Image} & \textbf{Question} & \textbf{Answer} \\
\midrule
\includegraphics[width=3cm]{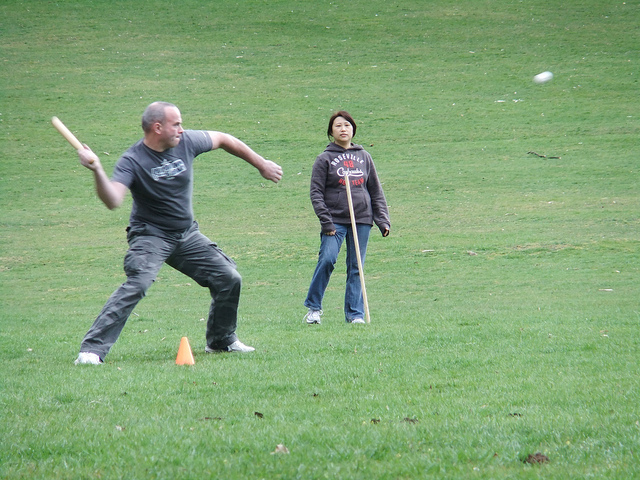} &
If you are the man in the image, where is the girl relative to your position? Left or Right? &
Ground-truth: Left \newline
GPT-4o: Right \xmark \newline
Gemini 2.0 Flash: Right \xmark \\

\includegraphics[width=3cm]{figures/appendix_examples/000000000063.jpg} &
From the perspective of the viewer, where is the girl relative to the man? Left or Right? &
Ground-truth: Right \newline
GPT-4o: Right \cmark \newline
Gemini 2.0 Flash: Right \cmark \\

\includegraphics[width=3cm]{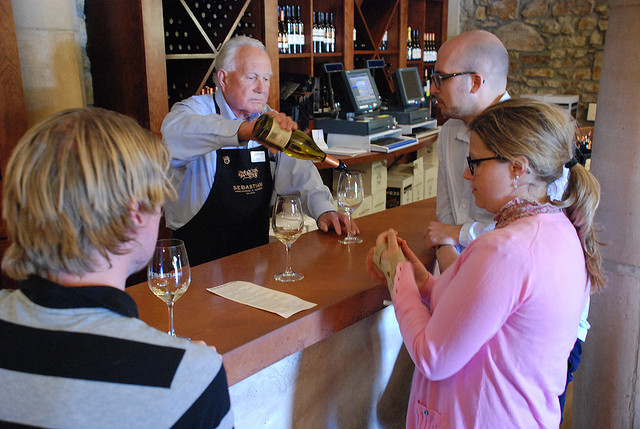} &
If you are the man pouring wine, where is the man with glasses relative to your position? Left or Right? &
Ground-truth: Left \newline
GPT-4o: Right \xmark \newline
Gemini 2.0 Flash: Right \xmark \\
\bottomrule
\end{tabular}
\end{adjustbox}
\caption{Task: Position}
\end{subtable}

%%% Counting
\begin{subtable}[t]{\textwidth}
\begin{adjustbox}{max width=\textwidth}
\begin{tabular}{m{3cm} m{6cm} m{6cm}}
\toprule
\textbf{Image} & \textbf{Question} & \textbf{Answer} \\
\midrule
\includegraphics[width=3cm]{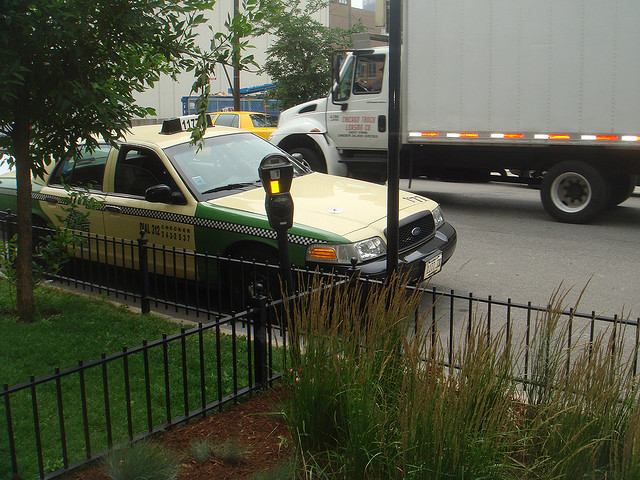} &
How many buses are visible? &
Ground-truth: 0 \newline
GPT-4o: 0 \cmark \newline
Gemini 2.0 Flash: 0 \cmark \\

\includegraphics[width=3cm]{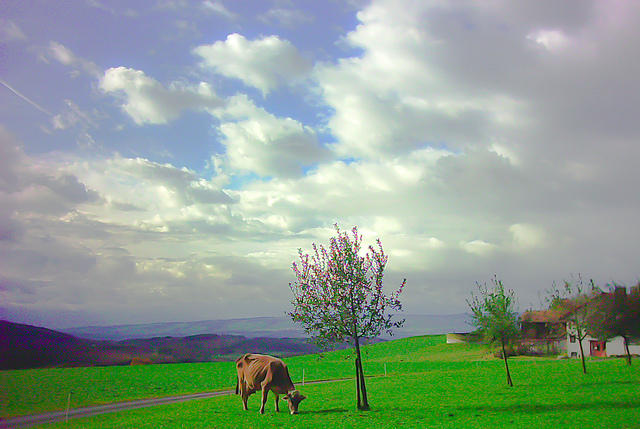} &
How many trees are visible? &
Ground-truth: 4 \newline
GPT-4o: 2 \xmark \newline
Gemini 2.0 Flash: 3 \xmark \\

\includegraphics[width=3cm]{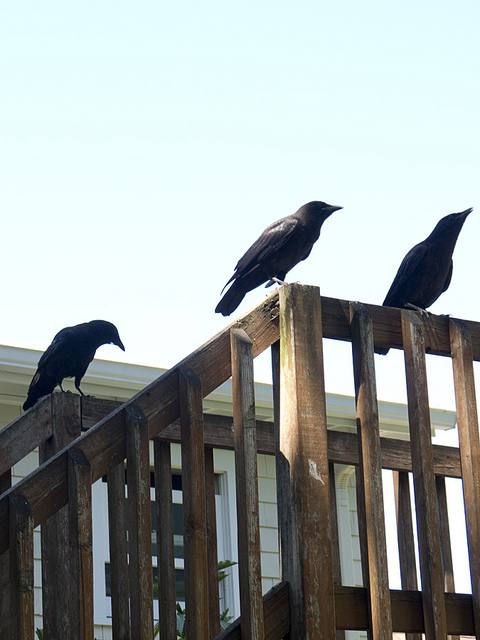} &
How many crows can be seen? &
Ground-truth: 3 \newline
GPT-4o: 3 \cmark \newline
Gemini 2.0 Flash: 3 \cmark \\
\bottomrule
\end{tabular}
\end{adjustbox}
\caption{Task: Counting}
\end{subtable}
\end{table*}

\begin{table*}[tb]
\centering
\ContinuedFloat

%%% Distance
\begin{subtable}[t]{\textwidth}
\begin{adjustbox}{max width=\textwidth}
\begin{tabular}{m{3cm} m{6cm} m{6cm}}
\toprule
\textbf{Image} & \textbf{Question} & \textbf{Answer} \\
\midrule
\includegraphics[width=3cm]{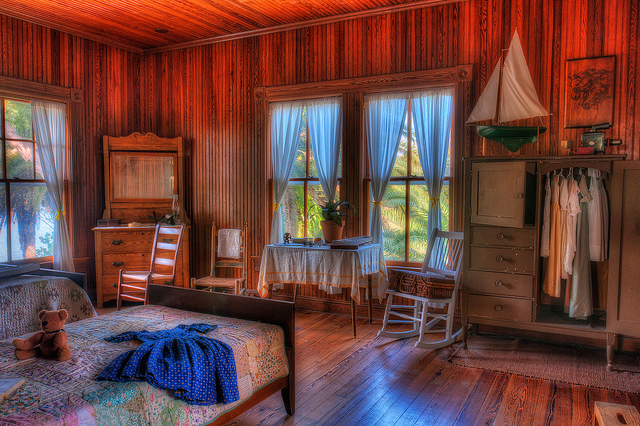} &
Which object is more distant from the plant? Chair or Table? &
Ground-truth: Chair \newline
GPT-4o: Chair \cmark \newline
Gemini 2.0 Flash: Chair \cmark \\

\includegraphics[width=3cm]{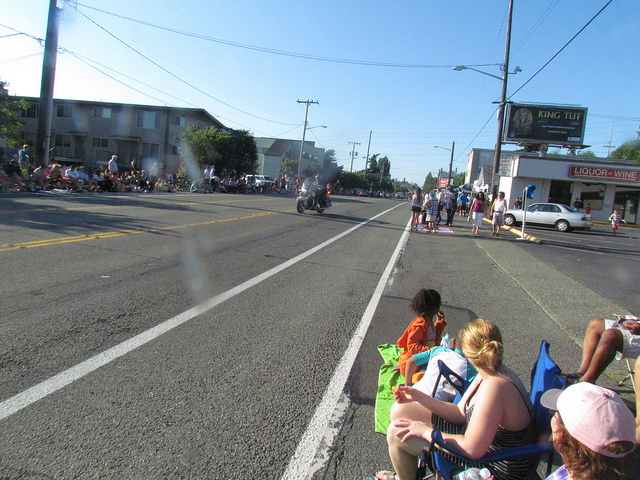} &
Which group of people is farther away from the motorcycle? People sitting on the left or People standing on the right? &
Ground-truth: People sitting on the left \newline
GPT-4o: People standing on the right \xmark \newline
Gemini 2.0 Flash: People sitting on the left \cmark \\

\includegraphics[width=3cm]{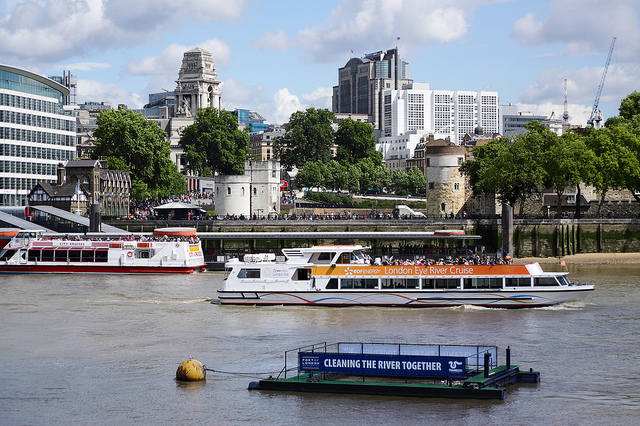} &
Which boat has a shorter distance to the bank? Boat with a cleaning sign or Boat with passengers? &
Ground-truth: Boat with passengers \newline
GPT-4o: Boat with a cleaning sign \xmark \newline
Gemini 2.0 Flash: Boat with a cleaning sign \xmark \\
\bottomrule
\end{tabular}
\end{adjustbox}
\caption{Task: Distance}
\end{subtable}

%%% Size
\begin{subtable}[t]{\textwidth}
\begin{adjustbox}{max width=\textwidth}
\begin{tabular}{m{3cm} m{6cm} m{6cm}}
\toprule
\textbf{Image} & \textbf{Question} & \textbf{Answer} \\
\midrule
\includegraphics[width=3cm]{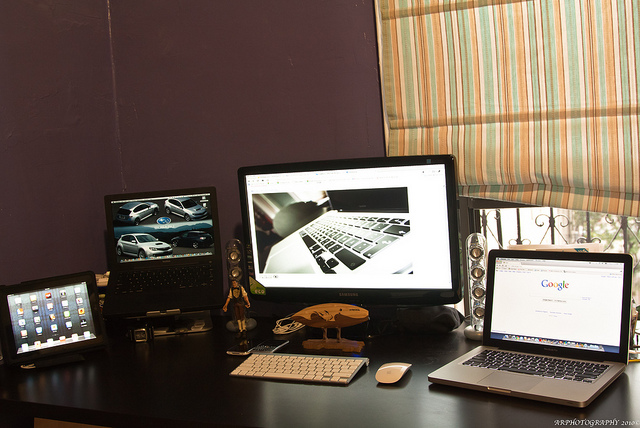} &
Which computer is the smallest? The leftmost one or The rightmost one? &
Ground-truth: The leftmost one \newline
GPT-4o: The leftmost one \cmark \newline
Gemini 2.0 Flash: The leftmost one \cmark \\

\includegraphics[width=3cm]{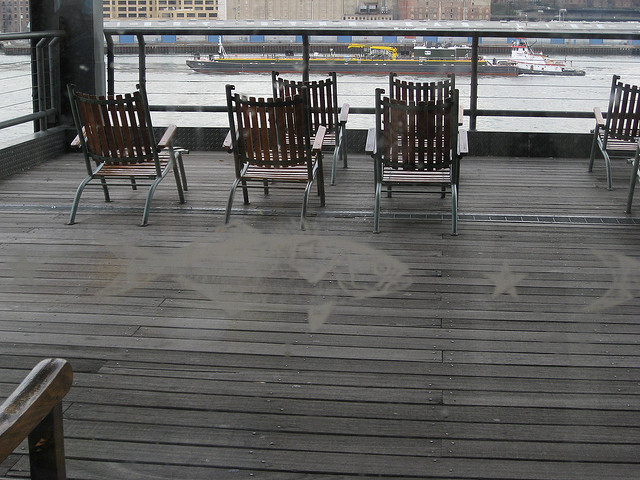} &
Is the fish icon on the ground bigger than the star icon in the picture? Yes or No? &
Ground-truth: Yes \newline
GPT-4o: Yes \cmark \newline
Gemini 2.0 Flash: Yes \cmark \\

\includegraphics[width=3cm]{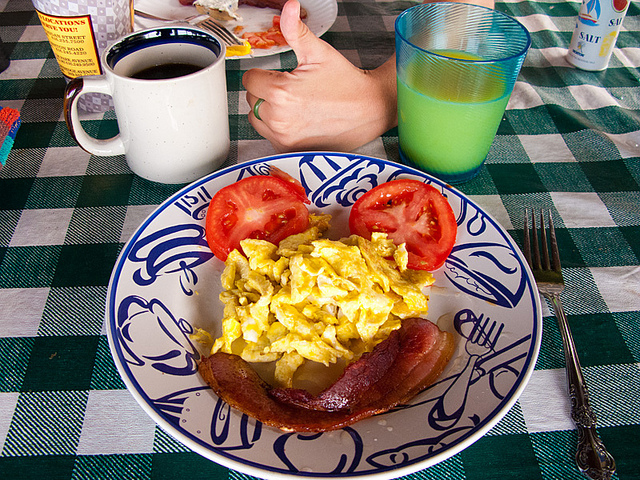} &
Which cup is higher? The left one or The right one? &
Ground-truth: The right one \newline
GPT-4o: The left one \xmark \newline
Gemini 2.0 Flash: The left one \xmark \\
\bottomrule
\end{tabular}
\end{adjustbox}
\caption{Task: Size}
\end{subtable}

\caption{Examples from single-skill tasks. \label{tab: single skill examples}}
\end{table*}
\begin{table*}[tb]
\centering

%%% Position + Counting
\begin{subtable}[t]{\textwidth}
\begin{adjustbox}{max width=\textwidth}
\begin{tabular}{m{3cm} m{6cm} m{6cm}}
\toprule
\textbf{Image} & \textbf{Question} & \textbf{Answer} \\
\midrule
\includegraphics[width=3cm]{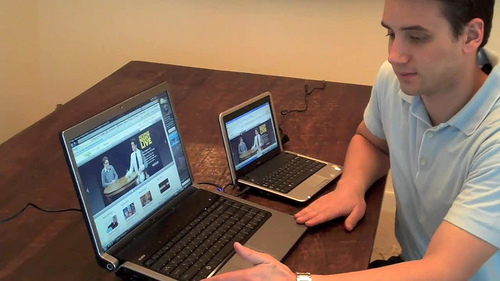} &
From the perspective of the viewer, how many laptops are on the left side of the man? &
Ground-truth: 2 \newline
GPT-4o: 2 \cmark \newline
Gemini 2.0 Flash: 1 \xmark \\

\includegraphics[width=3cm]{figures/appendix_examples/000000001459.jpg} &
If you are the man in the image, how many laptops are on your left? &
Ground-truth: 0 \newline
GPT-4o: 2 \xmark \newline
Gemini 2.0 Flash: 2 \xmark \\

\includegraphics[width=3cm]{figures/appendix_examples/000000000509.jpg} &
How many displays are to the right of the keyboard? &
Ground-truth: 1 \newline
GPT-4o: 2 \xmark \newline
Gemini 2.0 Flash: 3 \xmark \\
\bottomrule
\end{tabular}
\end{adjustbox}
\caption{Task: Position + Counting}
\end{subtable}

%%% Distance + Counting
\begin{subtable}[t]{\textwidth}
\begin{adjustbox}{max width=\textwidth}
\begin{tabular}{m{3cm} m{6cm} m{6cm}}
\toprule
\textbf{Image} & \textbf{Question} & \textbf{Answer} \\
\midrule
\includegraphics[width=3cm]{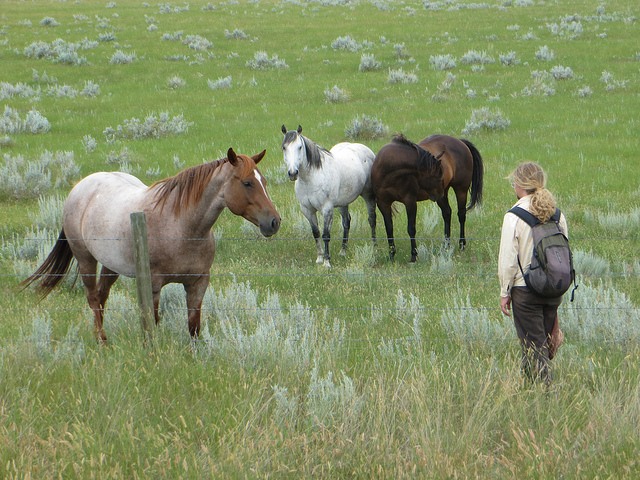} &
From the perspective of the person, how many horses are farther from him than the fence?  &
Ground-truth: 3 \newline
GPT-4o: 3 \cmark \newline
Gemini 2.0 Flash: 2 \xmark \\

\includegraphics[width=3cm]{figures/appendix_examples/000000554612.jpg} &
From the perspective of the person, how many horses are closer to him than the fence? &
Ground-truth: 0 \newline
GPT-4o: 1 \xmark \newline
Gemini 2.0 Flash: 1 \xmark \\

\includegraphics[width=3cm]{figures/appendix_examples/000000001385.jpg} &
How many crows are on the railing farther from the viewer? &
Ground-truth: 1 \newline
GPT-4o: 3 \xmark \newline
Gemini 2.0 Flash: 2 \xmark \\
\bottomrule
\end{tabular}
\end{adjustbox}
\caption{Task: Distance + Counting}
\end{subtable}
\end{table*}

\begin{table*}[tb]
\centering
\ContinuedFloat

%%% Distance + Size
\begin{subtable}[t]{\textwidth}
\begin{adjustbox}{max width=\textwidth}
\begin{tabular}{m{3cm} m{6cm} m{6cm}}
\toprule
\textbf{Image} & \textbf{Question} & \textbf{Answer} \\
\midrule
\includegraphics[width=3cm]{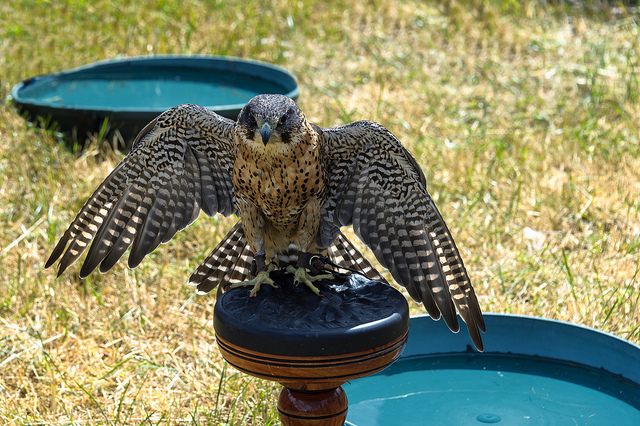} &
Are the turqoise plates likely the same size? Yes or No? &
Ground-truth: Yes \newline
GPT-4o: Yes \cmark \newline
Gemini 2.0 Flash: Yes \cmark \\

\includegraphics[width=3cm]{figures/appendix_examples/000000000718.jpg} &
The chairs on the top half of the picture are differently sized? Yes or No? &
Ground-truth: No \newline
GPT-4o: Yes \xmark \newline
Gemini 2.0 Flash: Yes \xmark \\

\includegraphics[width=3cm]{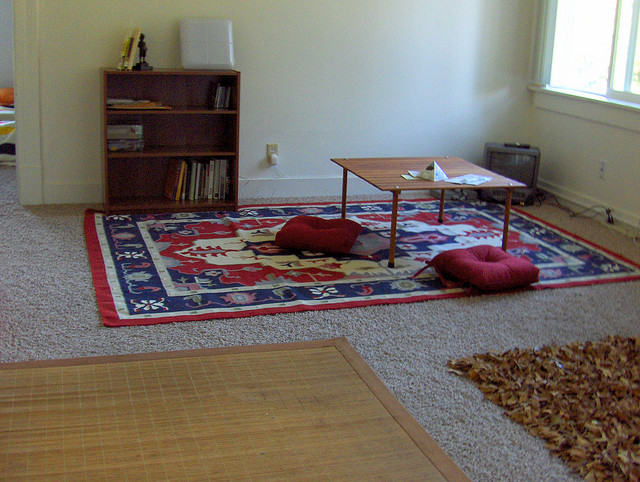} &
Which pillow is bigger? The left one or The right one or They are similar in size? &
Ground-truth: They are similar in size \newline
GPT-4o: The left one \xmark \newline
Gemini 2.0 Flash: The right one \xmark \\
\bottomrule
\end{tabular}
\end{adjustbox}
\caption{Task: Distance + Size}
\end{subtable}

\caption{Examples from multi-skill tasks. \label{tab: multi skill examples}}
\end{table*}
\begin{table*}[tb]
\centering

%%% Occlusion
\begin{subtable}[t]{\textwidth}
\begin{adjustbox}{max width=\textwidth}
\begin{tabular}{m{3cm} m{6cm} m{6cm}}
\toprule
\textbf{Image} & \textbf{Question} & \textbf{Answer} \\
\midrule
\includegraphics[width=3cm]{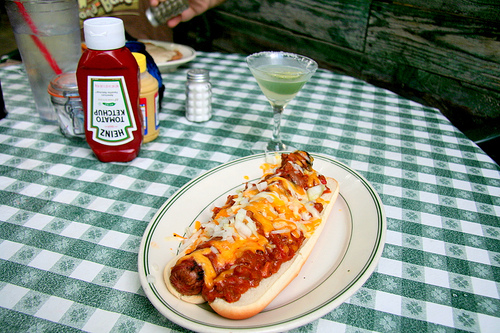} &
Where can the pepper shaker most likely be found? Behind the ketchup bottle or Behind the margarita glass? &
Ground-truth: Behind the ketchup bottle \newline
GPT-4o: Behind the ketchup bottle \cmark \newline
Gemini 2.0 Flash: Behind the ketchup bottle \cmark \\

\includegraphics[width=3cm]{figures/appendix_examples/000000000275.jpg} &
Where can another teddy bear be found? Inside the suitcase on the white chair or Under the blue dress? &
Ground-truth: Inside the suitcase on the white chair \newline
GPT-4o: Under the blue dress \xmark \newline
Gemini 2.0 Flash: Under the blue dress \xmark \\

\includegraphics[width=3cm]{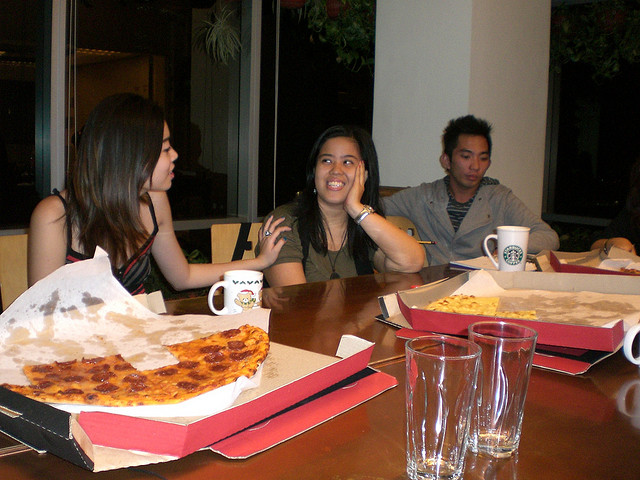} &
Is it likely that more glasses are hidden behind the pizza boxes? Yes or No? &
Ground-truth: No \newline
GPT-4o: No \cmark \newline
Gemini 2.0 Flash: Yes \xmark \\
\bottomrule
\end{tabular}
\end{adjustbox}
\caption{Task: Object occlusion}
\end{subtable}

%%% Manipulation
\begin{subtable}[t]{\textwidth}
\begin{adjustbox}{max width=\textwidth}
\begin{tabular}{m{3cm} m{6cm} m{6cm}}
\toprule
\textbf{Image} & \textbf{Question} & \textbf{Answer} \\
\midrule
\includegraphics[width=3cm]{figures/appendix_examples/000000000069.jpg} &
From the perspective of the womain in pink, can the man wearing glasses stand on her left side? Yes or No? &
Ground-truth: Yes \newline
GPT-4o: Yes \cmark \newline
Gemini 2.0 Flash: Yes \cmark \\

\includegraphics[width=3cm]{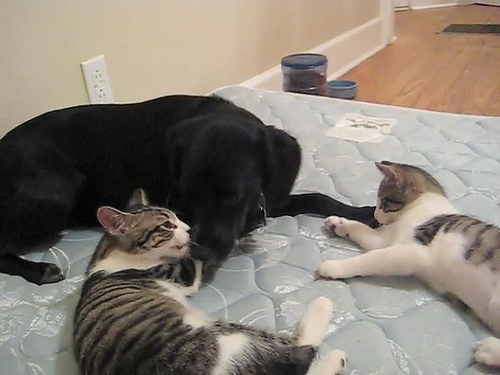} &
Can the dog lie between the two cats? Yes or No? &
Ground-truth: No \newline
GPT-4o: Yes \xmark \newline
Gemini 2.0 Flash: No \cmark \\

\includegraphics[width=3cm]{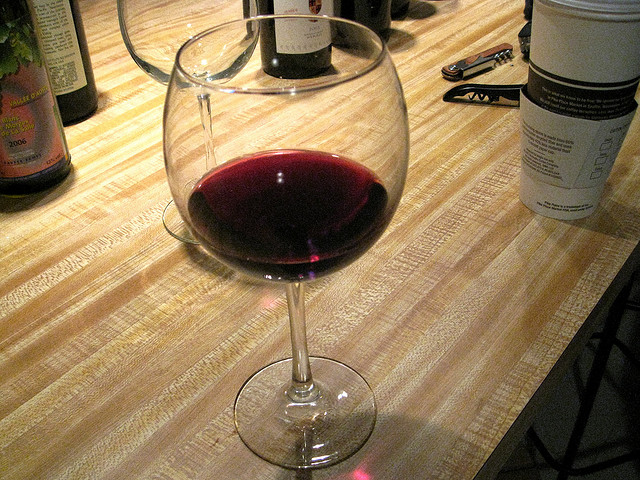} &
From the perspective of the viewer, can we pour all the wine from the nearer glass into the one behind? Yes or No? &
Ground-truth: Yes \newline
GPT-4o: Yes \cmark \newline
Gemini 2.0 Flash: No \xmark \\
\bottomrule
\end{tabular}
\end{adjustbox}
\caption{Task: Object manipulation}
\end{subtable}

\caption{Examples from reasoning tasks. \label{tab: reasoning examples}}
\end{table*}

\end{document}